%% file: main.tex
\documentclass[times]{article}

\usepackage{authblk}
\usepackage[numbers]{natbib}
\usepackage{geometry}
\input{preamble.tex}

\author[1]{Tyler Chang}
\author[2]{Andrew Gillette}
\author[1,3]{Romit Maulik}

\affil[1]{Mathematics and Computer Science Division, Argonne National Laboratory}
\affil[2]{Center for Applied Scientific Computing, Lawrence Livermore National Laboratory}
\affil[3]{Information Sciences and Technology Department, The Pennsylvania State University}

\title{Leveraging Interpolation Models and Error Bounds for Verifiable Scientific Machine Learning}

\begin{document}

\maketitle

\begin{abstract}
   Effective verification and validation techniques for modern scientific machine learning workflows are challenging to devise.  Statistical methods are abundant and easily deployed, but often rely on speculative assumptions about the data and methods involved.
   Error bounds for classical interpolation techniques can provide mathematically rigorous estimates of accuracy, but often are difficult or impractical to determine computationally.  In this work, we present a best-of-both-worlds approach to verifiable scientific machine learning by demonstrating that (1) multiple standard interpolation techniques have informative error bounds that can be computed or estimated efficiently;  (2) comparative performance among distinct interpolants can aid in validation goals;  (3) deploying interpolation methods on latent spaces generated by deep learning techniques enables some interpretability for black-box models.  We present a detailed case study of our approach for predicting lift-drag ratios from airfoil images.  Code developed for this work is available in a public Github repository.
\end{abstract}



\section{Introduction}

In recent years, data-driven surrogate modeling research has shown great promise in improving the predictability and efficiency of computational physics applications \cite{vinuesa2021potential,carleo2019machine}.
Surrogate models frequently utilize deep learning techniques for function approximation and have found application in control \cite{drgona2020differentiable}, optimization \cite{li2020efficient,sun2020surrogate}, uncertainty quantification \cite{tripathy2018deep,zhu2018bayesian,maulik2023quantifying}, and data assimilation tasks \cite{tang2020deep,maulik2022efficient}, among others.
However, the widespread adoption of deep learning models is still limited by their black-box nature, which makes their results difficult to interpret and verify \cite{rudin2019stop}.
Therefore, devising new models, techniques, and workflows that are interpretable and verifiable is considered one of the grand challenge problems for the next decade of scientific machine learning \cite{stevens2020ai}.

In this study, we focus on the importance of predictive modeling in the context of computational physics.
Specifically, we will analyze the usage of established interpolation models for surrogate modeling of complex physical systems, characterized by high computational costs when using first-principles based techniques.
We will address a common and essential problem formulation that arises in deep learning methods, machine-learning-based regression, and classical multivariate interpolation: given data from a multidimensional continuous input space, predict a single continuous response variable.
More precisely:
\[\begin{array}{l}
\text{given a training set ${\cal D}$ consisting of $n$ data points $\{x_i\}\subset\mathbb{R}^d$ paired with response values $\{f(x_i)\}\subset\mathbb{R}$,}\\ 
\text{predict the response values for previously unseen inputs within some pre-determined region ${\cal X}\subset\mathbb{R}^d$.}
\end{array}
\]

{
A sample application---which we will consider via a detailed case study later in the paper---is the prediction of the lift-to-drag ratio for an airfoil given only an image of its cross-section.
Training data in this context is a collection of lift-to-drag ratios for some airfoil designs, determined via computationally expensive numerical simulations of the Navier-Stokes equations.
Encoding the space of airfoil cross section images into a latent space $\mathbb{R}^d$ for some small $d$ is now standard machine learning practice via autoencoder methods.
We will show that analyzing the performance of standard interpolation techniques on the encoded images---i.e., from $\mathbb{R}^d$ to $\mathbb{R}^1$---reveals unique information about the validity of the trained autoencoder and enables simple visual interpretations of its failures.
}

One of the main advantages to restricting our study to interpolation as opposed to including regression, is that many interpolation methods have strict and easily-interpretable error bounds and well-known convergence rates \cite{cheneyandlight2009}.
In the context of scientific machine learning, these error bounds can be leveraged to {\sl verify} the correctness of a model and its applicability to a given problem.
Additionally, the direct dependence of these models and their bounds on the training data or simple basis functions, allows for predictions to be {\sl interpreted}, particularly when working with small- to medium-sized data sets.

In low- and medium-dimensional settings with noiseless (or low noise) training sets, interpolation techniques such as piecewise-linear interpolation, interpolating splines, and inverse-distance weightings have been shown to be competitive with machine learning techniques, such as multi-layer perceptrons (MLPs) and support vector regressors \cite{luxetal2018}.
Other interpolation techniques such as radial-basis function (RBF) models and Gaussian processes are frequently used as global \cite{muller2017,garnett2023} and local \cite{wildetal2008} surrogate models for optimization of physical systems.
By analyzing these methods and their error bounds and computing them on real scientific machine learning tasks, we will demonstrate the usefulness of these techniques for producing verifiable scientific machine learning workflows.
However, we must first address several challenges that have limited the usage of interpolation models for these regimes in the past.

First, regardless of the approximation techniques used, the predictive accuracy of the resulting model is fundamentally limited by the size and quality of ${\cal D}$, relative to the complexity of the problem.
In particular, the black-box problem formulation given in the previous paragraph is extremely sensitive to the {\sl curse of dimensionality}.
As the dimension $d$ of the problem increases, it is expected that maintaining a fixed level of accuracy in the approximation of $f$ across the entirety of ${\cal X}$ will require exponentially more data (i.e.~exponential growth in $n$).
For generic data science problems in extremely high-dimensional settings, it is worth noting that the top-performing deep learning models utilize trained or hand-crafted embedding layers, which are able to automatically extract lower-dimensional features for modeling when the input space can be effectively embedded by exploiting some problem-specific structure(s) \cite{bronstein2021geometric}.
In the context of computational physics, these embeddings may be inspired by some domain knowledge or physical structure of the problem \cite{bronstein2017geometric,gorban2010principal,doerner1999stable,li2022transformer}.
One key observation is that these embedding layers effectively reduce the problem into a continuous latent space, in which the fully-connected layers form a response surface resembling that of many classical approximation/interpolation methods.

Second, many classical interpolation techniques are intended for smooth noiseless problems, whereas real-world functions are often noisy or non-smooth.
It is often believed that (1) interpolation models will not perform well in these high-dimensional black-box modeling regimes, due to risk of {\sl overfitting} to non-smoothness and noise in the training data, and (2) machine learning models balance training loss against overfitting -- often referred to as the {\sl bias-variance tradeoff curve}.
However, contrary to conventional wisdom, recent work has shown that large over-parameterized models that interpolate or nearly-interpolate training data often achieve state-of-the-art performance on standard benchmark problems~\cite{zhangetal2021},
and interpolation techniques perform similarly to many of the machine learning techniques discussed above in high-dimensional regimes, so long as the signal-to-noise ratio is within reason~\cite{belkinetal2018,belkinetal2021,luxetal2021}.

In this paper, we propose a surrogate modeling strategy that utilizes interpolation methods together with their error bounds for verifiability, thereby alleviating a significant limitation of deep learning methods.
Our main contribution is an empirical study of how various known interpolation error bounds can be used to verify modern machine learning workflows.
To do so, we have proposed best practices and approximations for several well-known interpolation techniques and their error bounds in order make them computable for practical applications.
We also demonstrate how interpolation methods can be leveraged in a real-world scientific machine learning workflow that leverages deep learning techniques to extract low-dimensional representations of challenging high-dimensional problems.
Finally, a convenient aspect of certain interpolation methods is that each prediction is interpretable, allowing us to provide additional diagnostic information when validating results.

Our work is partially inspired by the paradigm of geometric deep learning \cite{bronstein2021geometric}, which suggests that all modern deep learning can be decomposed into representation learning vs.\ regression layers.
We are also inspired by the theoretical work of Belkin~\cite{belkinetal2021}, who has shown that interpolation is an effective paradigm for high-dimensional learning.
Finally, we build upon the recent work of Lux et al.~\cite{luxetal2021} who studied how interpolation methods can be effectively utilized for a wide variety of regression problems, and Gillette and Kur~\cite{gillette2023} who demonstrated how Delaunay interpolation convergence rates could be used to validate regression outputs.
Setting ourselves apart from these previous works, here, we consider error bounds for various distinct interpolation methods and observe the properties of a problem that make it amenable to prediction by each of these methods, in order to derive best practices.
Since there are many advantages to utilizing deep learning for scientific applications, {\bf our main motivation is to identify how deep learning could be augmented with interpolation, especially during model validation}.
We argue that validation is an ideal setting to utilize interpolation, since most interpolation methods do not require significant hyperparameter tuning and offer predictable performance for well-defined problems.

The remainder of this paper is organized as follows.
In Section~\ref{sec:interp-overview}, we will introduce several well-known interpolation methods and describe their relevant computations and error bounds.
We will also compare their pros and cons, specifically with respect to MLP-based regression models.
In Section~\ref{sec:empirical-study}, we will compare these methods for mathematically-defined test functions, with controllable smoothness, skew, and data spacing factors.
This allows us to assess the robustness of each method, and speculate about when each might be most useful.
In Section~\ref{sec:sciml-experiments}, we provide an example where these methods are used and compared on a scientific machine learning case study related to airfoil modeling, in order to demonstrate how these methods can be used in a scientific machine learning workflow that also leverages representation learning.
Finally, in Section~\ref{sec:conclusion}, we summarize our results and provide final recommendations about how to use interpolation for scientific machine learning verification.

\section{Interpolation Methods and Error Bounds}
\label{sec:interp-overview}

In this section we discuss common methods for interpolating an unknown blackbox function $f : \mathbb{R}^d \rightarrow \mathbb{R}$ {at an arbitrary ``test'' point given arbitrary training data.  More precisely, we assume we have access to} training data inputs ${\cal D} = \{x_i\}_{i=1}^n\subset\mathbb{R}^d$ and paired outputs $f\left(\cal D\right) = \{y_i\}_{i=1}^n\subset\mathbb{R}$, {with} an arbitrary ``test'' or ``query'' point $q\in{\cal X}\subset\mathbb{R}^d$.
We have chosen common methods from the literature that are well-defined for large values of $n$ and $d$, and do not rely upon any structured arrangement of ${\cal D}$.
All of our analysis is also based upon scattered training data ${\cal D}$, which is the least restrictive case.

We generally assume that $f$ is Lipschitz continuous, which is a standard assumption, but we do not assume any prior knowledge about the structure of $f$, such as gradient information, prior expectations, or constraints from physics.
For computational workflows where such knowledge is used or required, our results could be adopted from this general setting to a more specific one.
For some of our error analyses, additional assumptions about $f$ will be required, which we will highlight explicitly.

For each of the methods, we have organized the corresponding subsection into a mathematical summary of the method, followed by sub-subsections for a description of the error bounds, advantages and challenges of each technique, and description of our computational methods.
To this last point, for each technique, we have carefully chosen a specific method for calculating a prediction and its error bounds, in order to avoid ambiguity.
These techniques are chosen primarily to best reflect the theory, so that the error bounds will be most practical and the techniques can be used consistently across a wide variety of scientific machine learning problems.
In particular, we have intentionally avoided tuning any hyperparameters for specific problems, as this would introduce an additional source of error and bias, which would go against our motivation to use interpolation methods for verification of scientific machine learning workflows.

\subsection{Mesh-based Interpolation and the Delaunay Interpolant}
\label{sec:delaunay-interp}

In low dimensions, unstructured mesh-based interpolants are widely used for scattered data sets in computational geometry, visualization, and physical modeling.  
Error bounds for mesh-based interpolants are well-understood, especially in the context of finite element methods~\cite{shewchuk2002proceedings}.  
Few mesh-based methods can be extended to arbitrary dimensions, however, due to the so-called the ``curse of dimensionality'' for geometry-based methods.

Notably, piecewise linear interpolation with respect to a simplicial mesh structure can be carried out in a scalable and computationally efficient manner in higher dimensions.
To define a piecewise linear interpolant, one needs a simplicial mesh over the training data inputs, denoted $T({\cal D})$, such that every simplex in $T({\cal D})$ has vertices in ${\cal D}$.
Putting aside the construction of the mesh for a moment, suppose the query point $q$ is contained in the simplex $S^{(i)}_T \in T({\cal D})$, with vertices $\{x^{(i,0)}, \ldots, x^{(i,d)}\}$.
Then there exist unique interpolation weights $w = \{w_0, \ldots, w_d\}$ given by solving
\[
A^{(i)}_{T}w_{1:d} = q, \quad w_0 = 1 - \sum_{i=1}^d w_i
\]
where
\begin{equation}
\label{eq:interp-mat}
A^{(i)}_T = \left[ x^{(i,1)} - x^{(i,0)} \quad|\quad x^{(i,2)} - x^{(i,0)} \quad|\quad \ldots \quad|\quad x^{(i,d)} - x^{(i,0)} \right].
\end{equation}
The piecewise linear interpolant ${\hat f}_T$ for $q$ in $S^{(i)}_T$ is then defined as the weighted sum
\begin{equation}
\label{eq:delaunay-interp}
{\hat f}_{T}(q) := \sum_{j=0}^{d} w_j f\left(x^{(i,j)}\right)
\end{equation}
Note that it is possible for $q$ to be contained in the boundary between multiple simplices, but the value of ${\hat f}_T(q)$ will agree along these boundaries such that ${\hat f_T}$ is well-defined and continuous on the entire convex hull of $\cal D$.
This method is called barycentric interpolation and dates back to work by M\"obius~\cite{mobius1827}.

To define a mesh $T$,  we employ Delaunay triangulation~\cite{delaunay1934sphere}, a classical technique that is widely used for unstructured meshes in $d=2$ or $3$~\cite{chengetal2012}, but is defined without restriction on the dimension $d$.
The Delaunay triangulation $DT({\cal D})$ is a simplicial mesh whose elements are determined by the ``open circumball'' property.
Specifically, for each simplex $S^{(i)} \in DT({\cal D})$, the open circumball $B^{(i)}$ about $S^{(i)}$ satisfies $B^{(i)}\cap {\cal D} = \emptyset$.
For ${\cal D}$ in {\sl general position}, this definition is also sufficient to guarantee existence and uniqueness of $DT({\cal D})$.
We will use the notation ${\hat f_{DT}}$ to mean the the interpolant defined in (\ref{eq:delaunay-interp}) with mesh $T:=DT({\cal D})$.


\subsubsection{Error bounds for Delaunay and other simplex interpolants}
\label{sec:delaunay-bounds}

The generic piecewise linear interpolant ${\hat f}_T$ from (\ref{eq:delaunay-interp}) can be used to approximate any Lipschitz continuous $f$.
Under the stronger assumption that $f\in {\cal C}^1$ and every component of $\nabla f$ is $\gamma$-Lipschitz continuous (for some fixed $\gamma>0$), it is shown in~\cite{luxetal2021} that the following first-order Taylor-like bound holds:
\begin{equation}
\label{eq:delaunay-bound-1}
|f(q) - {\hat f}_T(q)| \leq \frac{\gamma\|q - x^{(i,0)}\|_2^2}{2} + \frac{\sqrt{d}\gamma k^2}{2\sigma_d} \|q - x^{(i,0)}\|_2.
\end{equation}
where $k = \max_{j\in\{1,\ldots,d\}}\|x^{(i,0)} - x^{(i,d)}\|_2$ and $\sigma_d$ is the smallest singular value of $A^{(i)}_T$.
Note that in practice, $x^{(i,0)}$ can be chosen to be the nearest vertex of $S^{(i)}_T$ to $q$.
We remark that a similar bound is derived in \cite{regis2015} under slightly different assumptions.

Let $h$ denote the maximum diameter of a simplex in the mesh.
Since both $\|q - x^{(i,0)}\|_2$ and $k$ are functions of the density of ${\cal D}$ in the query region ${\cal X}$, they are each bounded above by $h$.  
Accordingly, the above bound reduces to
\begin{equation}
\label{eq:delaunay-bound-2}
|f(q) - {\hat f}_T(q)| \leq \frac{\gamma}{2}h^2 + \frac{\sqrt{d}\gamma}{2}\frac{k}{\sigma_d} h^2.
\end{equation}
Notice that $k/\sigma_d$ can be upper-bounded by the condition number of $A^{(i)}_T$ and is actually an estimate for the aspect-ratio of the containing simplex.
In other words, so long as the aspect ratio of the simplices in $T({\cal D})$ can be upper-bounded, the interpolation error is $|f(q) - {\hat f}_T(q)| \in {\cal O}(h^2)$.
The correlation between good aspect ratio of simplices and good interpolation error estimates has led to various notions of optimality for Delaunay triangulations~\cite{chenandxu2004,rajan1994}.

\subsubsection{Advantages and challenges of Delaunay interpolants}
\label{sec:delaunay-challenges}

The Delaunay interpolant is entirely deterministic with no hyperparameters, a significant advantage 
over other types of interpolants that we will consider. 
There is no concept of {\sl training} or {\sl fitting} a Delaunay-based model since ${\hat f}_{DT}$ is well-defined. 
Consequently, there is a ``hidden'' savings in using Delaunay methods in that no time or effort needs to be invested in identifying suitable hyperparameters or training a model.
In addition, the simple form of ${\hat f}_{DT}$ provides a clear notion of interpretability: an interpolated value $q$ is uniquely determined by a weighted sum of the data points at the vertices of the simplex containing $q$.
Evidence for the accuracy and efficiency of Delaunay interpolants for regressing scientific data sets was shown in~\cite{luxetal2021}.
Additionally, even in the context of classification with noisy data, the risk of a Delaunay interpolation-based classifier has been shown to approach the Bayes risk in high dimensions \cite{belkinetal2018}.


A challenge when employing Delaunay interpolants is a potentially expensive computational cost at inference time.
Many common approaches for computing Delaunay triangulations are impractical in even moderate dimensions, due to the exponential size of $DT({\cal D})$ with growing $d$ \cite{klee1980}.
However, it is possible to identify the single simplex needed to calculate ${\hat f}_{DT}(q)$ for any given $q$ at inference time, for the cost of solving a linear program \cite{changetal2018b,fukuda2004}.
This approach is scalable enough for usage with moderately-sized training sets, but still considerably more expensive than making a single prediction with many comparable neural network models, such as MLPs.

Additionally, when implementing such an approach, one must be careful of geometric degeneracy, which results in a degenerate linear program and could cause failures in certain solvers.
These degeneracies occur with probability zero in theory {for randomly scattered data}, but are extremely common in real-world data sets.
For instance, if any subset of a training set is grid-aligned {or if all the data lies in a hyperplane of codimension at least 1}, the geometry of the data would be considered degenerate.
{These degeneracies can be handled robustly.  Grid-aligned data still has associated Delaunay interpolants, but there may be more than one valid choice.  A hyperplane containing all the data can be detected and a Delaunay interpolant can be defined over this hyperplane.}

Another challenge of Delaunay interpolation is the failure to natively handle extrapolation outside the convex hull of the training data, denoted $CH({\cal D})$.
Consider an extrapolation point $z \in {\cal X}$, but $z\not\in CH({\cal D})$.
One solution is to handle this by projecting each such $z$ onto $CH({\cal D})$ and interpolating the projection ${\hat z}$.
Assuming that $f$ is $L$-Lipschitz continuous and extending the error bound in (\ref{eq:delaunay-bound-1}), we have the extrapolation error estimate 
\begin{equation}
\label{eq:delaunay-extrap}
|f(z) - {\hat f}_T(z)| \leq \frac{\gamma\|{\hat z} - x^{(i,0)}\|_2^2}{2} + \frac{\sqrt{d}\gamma k^2}{2\sigma_d} \|{\hat z} - x^{(i,0)}\|_2 + L \|{\hat z} - z\|_2.
\end{equation}
This is a reasonable approach when the extrapolation residual $\|{\hat z} - z\|_2$ is relatively small, but can lead to extremely large errors when extrapolating far from the convex hull of the data.
However, it is worth noting that large extrapolation residuals correspond to out-of-sample predictions.
Therefore, it is expected that many methods would suffer when making such predictions, and the extrapolation residual can serve as a useful piece of diagnostic information for any methodology.

\subsubsection{Our methods for Delaunay interpolation}
\label{sec:delaunay-methods}

In this paper, we use the {\tt DelaunaySparse} software, which computes only the single simplex from $DT({\cal D})$ containing $q$, as is needed to compute the value of ${\hat f}_{DT}(q)$ \cite{changetal2020c}.
One advantage of {\tt DelaunaySparse} is that it is extremely robust and handles issues that arise when working with degenerate training sets (i.e., not in general position) for little additional expense beyond the linear programming formulation.
Additionally, {\tt DelaunaySparse} automatically detects when a query point $q$ is outside $CH({\cal D})$ and projects $q$ onto the convex hull, as described in Section~\ref{sec:delaunay-challenges}.

To compute the error bounds from Section~\ref{sec:delaunay-bounds}, we first note that for real-world problems, the bound (\ref{eq:delaunay-bound-2}) is often overly pessimistic as it is derived from a worst-case analysis.
In fact, it is a Taylor-like bound centered at $x^{(i,0)}$, based on the fact that the error in simplex-gradient $c$ could be at most $\big\|\nabla f(x^{(i,0)}) - c\big\|_{2} \leq \sqrt{d}  \gamma k^{2} / 2 \sigma_{d}$.
Thus, $\sigma_d$ represents the worst-case error when using the $d$ directional derivatives given by columns of $A_{T}$ to estimate the gradient, assuming that the maximum change in $\nabla f$ occurs in the direction of least information.
Further, the value of $\gamma$ is effectively the worst-case Lipschitz constant for $\nabla f$ over all of ${\cal X}$, which may not be representative of the {\sl local} variation in $\nabla f$, a more relevant quantity for predicting a value at $q$.

We modify (\ref{eq:delaunay-bound-2}) into a computable estimate that better represents average-case error by applying the following relaxations at a point $q$ contained in simplex $S^{(i)}_T$:
\begin{itemize}
    \item Replace $\gamma$ with $\hat\gamma$, the {\sl local} Lipschitz constant for $\nabla f$ over $S^{(i)}_T$.  Estimate $\hat\gamma$ as the max over all three-point divided difference estimates using only vertices from $S^{(i)}_T$ .
    This corresponds to a relaxation that the componentwise Lipschitz constants of $\nabla f$ may change across the domain ${\cal X}$.
    \item Replace $\sigma_d$ with $\hat\sigma_d$, the {\sl average} over all singular-values of the matrix $A_T$ from (\ref{eq:interp-mat}).
    This corresponds to a relaxation that the maximum change in $\nabla f$ is typically not perfectly aligned with the direction of least information in $A_T$.
\end{itemize}
Using the above estimates, the revised bound for simplex-based piecewise linear interpolation becomes
\begin{equation}
\label{eq:delaunay-bound-rev}
|f(q) - {\hat f}_T(q)| \leq \frac{\hat\gamma}{2}h^2 + \frac{\sqrt{d}{\hat \gamma}}{2}\frac{k}{\hat\sigma_d} h^2.
\end{equation}
Note that this is no longer a worst-case bound, meaning it can be violated in practice.
However, we have observed empirically that the revised bound is more practical than the original for predicting the true error.

\subsection{RBF Interpolants and Thin-Plate Splines}
\label{sec:rbf-interp}

Radial-basis function (RBF) interpolants are families of interpolants that are defined by linear combinations of radially-symmetric nonlinear basis functions.
Given $n$ data points, the RBF interpolant requires a collection of $n$ basis functions, denoted $\Phi := \{\phi^{(i)}\}_{i=1}^n$.
Typically, each function $\phi_i$ is centered at a corresponding training point $x^{(i)} \in {\cal D}$ such that $\phi^{(i)}(q) = b_\Phi(\|q - x^{(i)}\|_2)$, where $b_\Phi$ is a one-dimensional basis function, fixed over all $\Phi$.
The function $b_\Phi: \mathbb{R}^1\rightarrow\mathbb{R}^1$ is called the \textit{kernel function} of the interpolant.

The first step in fitting an RBF interpolant is (typically) fitting a ``polynomial tail'' $\tail(q)$, computed via generalized linear regression with polynomial basis functions.
In this paper, we will use either a least-squares approximation for a linear tail, resulting in $\tail(q) = \tail_1(q) := \alpha^\top q + \beta$ for some $\alpha,\beta$, or a constant tail given by $\tail(q) = \tail_0(q) := \sum_{i=1}^n f(x^{(i)}) / n$.
{It is possible to employ higher order techniques in some settings, such as RBF-FD approximations~\cite{flyer2016role}, but such methods are only feasible in low dimensions, typically $d\leq3$.}

The next step is to solve the linear system given by
\begin{equation}
\label{eq:rbf-interp-system}
A_{\Phi} c = \left[
\begin{array}{c}
f(x^{(1)}) - \tail(x^{(1)})\\
\vdots \\
f(x^{(n)}) - \tail(x^{(n)})
\end{array} \right]
\end{equation}
where
\begin{equation}
\label{eq:rbf-matrix}
A_{\Phi} := \left[
\begin{array}{cccc}
\phi^{(1)}(x^{(1)}) & \phi^{(2)}(x^{(1)}) & \ldots & \phi^{(n)}(x^{(1)})\\
\phi^{(1)}(x^{(2)}) & \phi^{(2)}(x^{(2)}) & \ldots & \phi^{(n)}(x^{(2)})\\
\vdots      & \vdots      &        & \vdots\\
\phi^{(1)}(x^{(n)}) & \phi^{(2)}(x^{(n)}) & \ldots & \phi^{(n)}(x^{(n)})
\end{array}\right].
\end{equation}
The RBF interpolant ${\hat f}_\Phi$ for $q$ is then defined by
\begin{equation}
    \label{eq:rbf-interp-def}
    {\hat f}_\Phi(q) := \sum_{i=1}^n c_i\phi^{(i)}(q) + \tail(q).
\end{equation}

Note that since each $\phi^{(i)}$ is radially symmetric about $x^{(i)}$, $A_\Phi$ is symmetric.
When $\Phi$ are also positive definite, then $A_\Phi$ is symmetric positive definite, and (\ref{eq:rbf-interp-system}) can be solved via Cholesky decomposition, which is slightly faster than LU or QR decomposition.
However, in both cases the time complexity is cubic and space complexity is quadratic in $n$ \cite{golubandvanloan2013}.

\subsubsection{Error bounds for RBFs}
\label{sec:rbf-error}

Error analysis for RBF interpolation is tricky because the error bounds depend on globally-dependent quantities derived from $f$, which would rarely be known in practice.  
We focus on a bound provided by Powell \cite{powell1994uniform} and explained in \cite[Theorem 1]{buhmann_2000} as follows.
Define
\[h := \sup_{x\in\cal{X}}\inf_{x_i\in \cal{D}}\|x-x_i\|,\]
i.e., $h$ is the largest distance to a point in $\cal D$ over $\cal{X}$.
Then, for any neighborhood $\cal{X}\ni$ $q$ such that $0<h<1$, there is a constant $C$, independent of $h$, $\cal{D}$ and $q$, such that
\begin{equation}
\label{eq:rbf-est-gen}
|f(q) - {\hat f}_{\Phi}(q)| \leq C \|f\|_{\Phi} h\sqrt{\log(h^{-1})}
\end{equation}
where $\|f\|_{\Phi}$ is the homogeneous Sobolev norm of order 2 on $f$.\footnote{The norm on $f$ comes from the norm induced by the reproducing kernel Hilbert space for $f$, sometimes called the ``native space'' of $f$, which reduces to a homogeneous Sobolev norm in this case (by completion of native space with respect to the Sobolev seminorm); see~\cite{buhmann_2000} for more details.}
With additional higher order Sobolev or Lipschitz smoothness assumptions on $f$, the above bound can be improved to $\mathcal{O}(h^{2k+\eta})$ for some $\eta>0$ and then used to prove convergence results; see~\cite{buhmann_2000}.

Sticking with a mere Lipschitiz continuity assumption for $f$, we are not aware of a computable estimate for $C$ or $\|f\|_{\Phi}$.
However, it is useful to note the scaling behavior of (\ref{eq:rbf-est-gen}) with respect to $h$.
Note that $h$ is a measure of the density of  ${\cal D}$ in ${\cal X}$.
If ${\cal D}$ becomes dense enough, the estimate is dominated by the $\log(h^{-1})$ term, which blows up as $h\rightarrow 0$, in that large $h$ values indicate less density, {somewhere} in $\mathcal{X}$.
The blow up is a feature, not a bug, of the estimate, as it reflects the numerical ill-conditioning of the linear system  (\ref{eq:rbf-matrix}) as the training data clusters.

The behavior of the estimate highlights the so-called ``uncertainty principle'' for RBFs, which roughly states that it is not possible to achieve uniform convergence and solvability of ${\hat f}_\Phi$ \cite{schaback1995}.
This ill-conditioning is still particularly prevalent for unbalanced training sets, and may be amplified depending on the choice of basis function $\Phi$.

\subsubsection{Advantages and challenges of RBFs}
\label{sec:rbf-challenges}

When the size of the data set $n$ is small, RBF interpolants are relatively cheap to fit.
Additionally, {for} many choices of basis functions---including the thin-plate spline functions that we will describe in the next section---there are often no hyperparameters.
Evaluation of the interpolant at query-time is also quick with $O(nd)$ complexity.
It is thus no surprise that RBF methods are a popular off-the-shelf solution to a wide variety of interpolation problems.

As $n$ gets large, the complexity of solving (\ref{eq:rbf-interp-system}) becomes prohibitive.
The factorization of (\ref{eq:rbf-matrix}) has a time complexity ${\cal O}(n^3)$ and space complexity of ${\cal O}(n^2)$.
To reduce this burden for large problems, a hyperparameter $k$ is introduced and the interpolated value is determined using only the $k$ nearest neighbors.
However, cutting off based on a neighborhood criterion creates discontinuities in ${\hat f}_{RBF}$, which can be undesirable.
For the problems that we are working with, the data sets do not grow large enough in $n$ for the ${\cal O}(n^3)$ complexity to become a concern, so we do not consider any of the above approaches.
However, it is worthwhile to note these issues for larger problems.

For many common RBF kernels, $\Phi(x) \rightarrow 0$ as $x\rightarrow \infty$ so that each $\phi_i$ can be interpreted as a similarity metric with $x^{(i)}$, for $i=1$, $\ldots$, $n$.
However, this also implies that when $q$ is outside $CH({\cal X})$, then the predicted value of the RBF kernel will approach the value of the tail, $\tail(q)$.
This may be extremely inaccurate if $f$ is not well-approximated by a constant or linear function.
For other common RBF kernels, $\Phi(x)$ may diverge at infinity, in which case it is difficult to interpret the value of each $\phi_i(q)$, and the overall approximation ${\hat f}_{\Phi}(q)$ may also diverge when $q$ is outside $CH({\cal X})$.

Compared to the other interpolants that we consider, one critical challenge for a generic RBF interpolant is that the error bound cannot be easily calculated or approximated.
One critical unknown in (\ref{eq:rbf-est-gen}) is the value of the constants $C$ and $\|f\|_\Phi$.
In principle, these values can be determined independently of the data set ${\cal D}$, however in practice they cannot be known for many real-world applications.
We will see a special case in Section~\ref{sec:gp-bounds} where an error bound can be derived by statistical means, but this approach is kernel-dependent.

\subsubsection{Our methods for RBF interpolation and the TPS kernel}
\label{sec:rbf-methods}

In this paper, we focus on the thin-plate spline (TPS) kernel RBF interpolant.
The tail function is linear, i.e.~$\tail=\tail_1$, as defined in Section~\ref{sec:rbf-interp}.
The basis functions are $\phi^{(i)}(q) := \|q - x^{(i)}\|^2\log \|q - x^{(i)}\|$.
These choices are the default settings for the {\tt RBFInterpolator} class in {\tt scipy.interpolate} \cite{virtanenetal2020}, which we use in this rest of this paper.
We denote the resulting interpolant by ${\hat f}_{\text{TPS}}$.

We have selected to examine TPS RBF interpolants for several reasons.
First, TPS RBFs are considered ``standard,'' as evidenced by their role as the default setting for {\tt scipy.interpolate}.
Second, TPS RBFs are less sensitive to the conditioning issue noted in Section~\ref{sec:rbf-error} than other choices for basis functions $\Phi$, as we will see in Section~\ref{sec:gp-interp}.
Finally, it is worth noting that the TPS kernel has no additional hyperparameters, making it a convenient choice for verifying machine learning workflows.

However, TPS RBFs also face some drawbacks relative to other RBF kernels.
First, the TPS kernel $x^2\log|x|$ is not monotone decreasing, which prevents it from being interpreted as a similarity metric.
Accordingly, it can be difficult to interpret the predictions of a computed TPS RBF interpolant on a given dataset.
Further, since the kernel function diverges at infinity, the approximation of a TPS interpolant can become arbitrarily bad if the query point $q$ corresponds to geometric extrapolation, as described in Section~\ref{sec:rbf-error}.

For the error bounds, as noted in Section~\ref{sec:rbf-error}, it is not possible for us to know the true bound due to the constants $C$ and $\|f\|_\Phi$.
In this paper, we approximate the product of these constants by the Lipschitz constant $L$.
This yields a practical bound of 
\begin{equation}
\label{eq:RBF-practical}
|f(q) - {\hat f}_{TPS}| \leq L h \sqrt{\log h}.
\end{equation}
This bound is similar to a simple Lipschitz bound, which would be a reasonable scale and also maintains the same convergence rate as (\ref{eq:rbf-est-gen}).
Note that although there is no direct dependence on the dimension $d$, both $L$ and $h$ tend to increase with the dimension in practice.

\subsection{Gaussian Processes for Interpolation}
\label{sec:gp-interp}

Gaussian processes (GPs) are often motivated and used as a regression tool for contexts where approximation of a function is iteratively updated by the addition of new data.
In this work, since our data set is fixed and we are interested in interpolation rather than approximation, we find it more relevant to treat GPs as a special case of RBFs.~\footnote{Interpolatory GPs are closely related to Kriging models, commonly used in geostatistics; see~\cite{christianson2022traditional}.}
Following the notation from (\ref{eq:rbf-interp-system}), (\ref{eq:rbf-matrix}), and (\ref{eq:rbf-interp-def}), the GP kernel, denoted $\Phi=GP$, has Gaussian basis functions given by
\[\phi^{(i)}(q) = e^{-\|q-x^{(i)}\|^2/\tau},\]
where $\tau$ is a shape parameter that corresponds to the standard deviation of the spatial correlation between training points.

In many applications of GPs, one uses a domain-specific {\sl prior} function in place of the polynomial tail $\tail(q)$ from (\ref{eq:rbf-interp-def}).
Here, we do not assume access to any such prior and hence take the constant approximation $\tail(q) = \tail_0(q) := \sum_{i=1}^n f(x^{(i)}) / n$, which is standard in the GP literature for dealing with a ``black box'' function.
We denote the resulting interpolant by ${\hat f}_{\text{GP}}$.

\subsubsection{Approximate error bounds for GPs}
\label{sec:gp-bounds}

Since the type of GP we are considering is a special case of RBFs, deriving practical, guaranteed error bounds by direct analysis encounters the same challenges described in Section~\ref{sec:rbf-error}.
However, unlike the TPS method, GPs have a statistical interpretation  that admits computable uncertainty estimates based on the standard deviation of the posterior distribution.

We view  ${\hat f}_{GP}$ as the posterior distribution over all possible realizations of $f$ given the observed training data ${\cal D}$.
By this interpretation, we attain an easily computed uncertainty estimate for any value predicted by ${\hat f}_{GP}$.
Since the standard assumptions for a GP lead to a Gaussian posterior, two standard deviations above and below the mean yield (approximately) a 95\% confidence interval.
Thus, we take two standard deviations above and below the mean as approximate error bounds for ${\hat f}_{GP}$.

{In light of these assumptions}, following the analysis in \cite[Ch.~2]{garnett2023}, the posterior distribution for ${\hat f}_{GP}$ at $q$ is given by
\[
    p\left({\hat f}_{GP}(q)~|~{\cal D}\right) \sim {\cal N}\left(\mathbb{E}\left[{\hat f}_{GP}\right](q), K(q, x)\right),
\]
with variance
\[
   K(q, x) = \tau_f \left(e^{-\|q-x\|_2^2/\tau} - \left(\phi^{(1)}(q), \ldots, \phi^{(n)}(q)\right) A_{GP}^{-1}\left[\begin{array}{c}\phi^{(1)}(x)\\ \vdots \\ \phi^{(n)}(x)\end{array}\right]\right),
\]
where $\tau_f$ is {some estimate for the width of the range of $f$} and
$A_{GP}$ is given by the RBF interpolation matrix from (\ref{eq:rbf-matrix}) with $\Phi=GP$.
For an estimate of the variance at $q$, which can be used to evaluate the expected variance in $|f(q) - {\hat f}_{GP}(q)|$, this reduces to
\begin{equation}
    \label{eq:gp-var}
    K(q,q) = \tau_f\left(1 -  \left\|\left(\phi^{(1)}(q), \ldots, \phi^{(n)}(q)\right)\right\|_{A^{-1}_{GP}}^2 \right).
\end{equation}
Here, $\| \cdot \|_{A^{-1}_{GP}}$ is the 2-norm rescaled by $A^{-1}_{GP}$, which is also a norm since $A^{-1}_{GP}$ is symmetric positive definite.

Note that this interpretation of GPs relies on additional assumptions about the smoothness of $f$ that may not hold true in practice.
For many real-world $f$, it is incorrect to assume that response values are Gaussian distributions and correlated with other response values based on the exponential of the inverse squared distance in the input space.
{It is for this reason that $K(q,q)$ is only an estimate for the actual variance.}
Further, if $f$ is not smooth enough, the error approximation by the above interpretation yields poor estimates.
Nevertheless, the uncertainty estimate is widely used and is considered a useful heuristic in many applications{~\cite{gramacy2020surrogates}}.

\subsubsection{Advantages and challenges of GPs}
\label{sec:gp-challenges}

Since they are a form of RBF interpolants, the computational complexity of GPs is the same as previously described in Section~\ref{sec:rbf-challenges}, and we will not discuss them again here.
Aside from the TPS RBFs that were used in Section~\ref{sec:rbf-interp}, GPs have a number of recognized advantages, which explains their popularity.
First, the statistical interpretation of GPs enables the often useful uncertainty estimates described in Section~\ref{sec:gp-bounds}, an interpretation not available for TPS and many other RBF schemes.
Second, the  GP kernel can be interpreted as a similarity metric, which allows for interpretability of the predictions of the method, as discussed in Section~\ref{sec:rbf-challenges}.
Third, when used with a mean prior (i.e., constant tail $\tail_0$ from Section~\ref{sec:rbf-interp}) and applied well outside the convex hull of the training data, GPs give predictions that trend toward the mean of the training data with extremely high uncertainty estimates.
This is a reasonable and conservative approach to handle extrapolation compared to the possible divergence of the TPS RBF, but may still yield extremely large errors.

A major challenge when using the GP uncertainty estimate is that it requires numerous additional assumptions that may not hold-up in practice.
In particular, GPs have very strong smoothness assumptions about $f$ and assume uniform spatial correlation between response values across the entire query region ${\cal X}$.
In many scientific applications, these additional assumptions are likely too strong, causing the estimates to lose statistical accuracy.

Additionally, the Gaussian kernel introduces a shape parameter $\tau$, whose value greatly impacts the predictive performance of the GP model.
Recall the ``uncertainty principle'' for RBF interpolants discussed in Section~\ref{sec:rbf-error}.
This phenomenon is particularly bad in the case of GPs since $\sigma_n$ decays squared-exponentially with the pairwise distance between training points.
An alternative approach is to add a diagonal ``nugget'' matrix $\varepsilon I$ to $A_{GP}$, as is often done for GP {\sl regression}.
Proponents argue that this approach improves the support of the resulting interpolant while also dodging any numerical issues resulting from poor conditioning of $A_{GP}$ \cite{gramacyandlee2012}.
Since such an approach sacrifices the interpolation property and our study is focused on interpolation methods, we do not employ it here.

In order to prevent these matrices from reaching numerical singularity, in practice the shape parameter $\tau$ must be decreased to improve the condition number.
This in turn restricts the support of each kernel, thereby creating gaps in the support of ${\hat f}_{GP}$, and consequently greatly {\sl increasing} the global approximation error.
In order to fit a model that is both accurate and numerically stable, the RBF literature recommends choosing a shape parameter $\tau$ that is as large as possible without causing singularity \cite{schaback1995}.
In modern applications, significant time and energy can be required to perform hyperparameter tuning to find an appropriate value of $\tau$ for each function $f$ and data set ${\cal X}$.
Such \textit{ad hoc} methods interfere with our ability to use GP interpolants for verifying machine learning workflows and are thus avoided.

\subsubsection{Our methods for fitting GP interpolants}
\label{sec:gp-methods}

Compared to previous interpolation methods in Sections~\ref{sec:delaunay-methods} and \ref{sec:rbf-methods} where we also had to adjust the error formulae in order to compute them, the GP's confidence interval computations are clearly laid out in Section~\ref{sec:gp-bounds} without ambiguity.
Therefore, the main source of ambiguity is to determine the shape hyperparameter $\tau$ of the GP model.

In the GP modeling literature, authors often experiment with variations of the Gaussian kernel (many of which introduce additional hyperparameters) and identify suitable hyperparameters (including $\tau$) through experimentation and/or domain knowledge.
Often, this level of careful hyperparameter tuning is necessary to achieve state-of-the-art performance with GP models.
In this work, we specifically avoid such methods since they require further verification.

Instead, we solve a well-defined optimization problem based on the statistical literature to select a maximum likelihood estimate of $\tau$ via {\tt scikit-learn} \cite{scikit-learn}.
Specifically, the log-marginal likelihood for observing the response values $\{y_i\}_{i=1}^n$ given the data $\{x_i\}_{i=1}^n$ is
\[
\log p(y_i | x_i) = -\frac{1}{2}\sum_{i=1}^n y_i c_i - \sum_{i=1}^n \log L_{ii} - \frac{n}{2} \log 2\pi~{= \frac 12\log\det\left(A_{GP} \right),}
\]
where $L$ is a Cholesky factor of $A_{GP}$ \cite{rasmussen2006}.
{Recall that $c:=A^{-1}_{GP} f$ is a function of $\tau$ since  $A_{GP}$ is a function of $\tau$.}
In {\tt scikit-learn}, this marginal likelihood function is maximized over possible values of $\tau$ using a multi-start L-BFGS solver.
While this is still technically a hyperparameter optimization problem, the problem is relatively inexpensive to solve and extremely {reproducible} since we are tuning a single variable using second-order gradient-based methods.  

\subsection{Artificial Neural Network and Multilayer Perceptron Models}
\label{sec:nn-interp}

In recent years, artificial neural network (ANN) models have become popular methods for surrogate modeling in scientific machine learning \cite{baldi2018deep,choudhary2022recent,kovachkietal2021,kutz2017deep,raissietal2019,reichstein2019deep}.
Most of these architectures are regression-based and thus not, strictly speaking, interpolants.  
However, given that part of our motivation is to study the validation of ANNs for scientific machine learning via interpolation, it makes sense to consider some common ANN architectures here.
As previously outlined, this motivation is backed by theoretical evidence that it is safe to interpolate in modern complex applications \cite{belkinetal2021} and empirical evidence that many top-performing ANN models are, in effect, interpolating data~\cite{zhangetal2021}.

Following the terminology of the geometric deep learning paradigm \cite{bronstein2021geometric}, a modern scientific machine learning ANN architecture exploits domain knowledge via symmetries that enable it to ``learn'' lower-dimensional feature representations.
It then regresses these features with fully-connected layers, which may resemble traditional MLPs.
It is worth noting that many architectures do not make a clear distinction between layers that perform these two tasks and train weights for all kinds of layers together via backpropogation.
However, only these fully connected layers are comparable to interpolants.
Therefore, our study focuses on comparison between true interpolants and MLPs trained to (nearly) zero training error.

Our scope does not limit us from considering more complex scientific machine learning workflows, as will be demonstrated in Section~\ref{sec:sciml-experiments}.
In general, we will consider ANNs of the following form
\begin{equation}
    \label{eq:nn-model}
    {
    {\hat f}_{ANN} = {\hat f}_{MLP}\circ{\hat f}_R
    }
    \end{equation}
where ${\hat f}_R$ could be defined over a complex or higher-dimensional feature space but maps to a compact ${\cal X} \subset \mathbb{R}^d$, and ${\hat f}_{MLP}$ is a function from $\mathbb{R}^d \rightarrow \mathbb{R}$ accurate on some bounded query domain ${\cal X}$, as described in Section~\ref{sec:interp-overview}.
The encoder portion of a latent space autoencoder network is an example of a function ${\hat f}_R$ that fits the above description.
We will consider ${\hat f}_{MLP}$ of the form
\begin{equation}
    \label{eq:mlp-model}
    {\hat f}_{MLP}(x) = \ell^{(N)} \circ \ell^{(N-1)} \circ \ldots \circ \ell^{(0)}(x)
\end{equation}
where each ``layer'' $\ell^{(i)}$ is of the form
$$
\ell^{(i)}(x^{(i)}) = u(W^{(i)}x^{(i)} + b^{(i)})
$$
where $u(\cdot)$ is a nonlinear activation function, $W^{(i)}$ is a tensor {whose} dimensions are compatible with $x^{(i)}$, and $b^{(i)}$ is a tensor whose dimension match $W^{(i)}x^{(i)}$.

The values of $W^{(i)}$ and $b^{(i)}$ for all $\ell^{(i)}$ and any parameters in ${\hat f}_R$ are determined by applying gradient descent (training) with a user-provided loss function via reverse-mode algorithmic differentiation (backpropagation).

\subsubsection{Advantages and Challenges}
\label{sec:nn-challenges}

One of the driving advantages of neural networks is that they are extremely flexible to utilizing different representation learning schemes (functions ${\hat f}_R$) and large networks can be trained in a scalable fashion with extremely large datasets.
In the context of scientific machine learning, the function ${\hat f}_R$ and/or the loss function can be customized to incorporate scientific domain knowledge and exploit symmetries in the problem formulation \cite{battaglia2018relational,goyal2022inductive}.

Another practical advantage is that while ANN models may require significant amounts of time and computational resources for training, they are generally much cheaper to evaluate at inference time, which makes them easy to deploy.
Additionally, compared to the interpolation models, they do not require access to the training set at inference time, which could have privacy advantages in certain contexts.
For these reasons, it is likely impractical to fully replace ANN models (or even portions of them) in many real-world applications.

However, the major drawbacks of ANN models is that given the formulation in (\ref{eq:nn-model}) and trained model weights, it is extremely difficult to interpret {\sl why} a network made a given prediction.
This coupled with the approach to selecting parameter weights and lack of any hard error bounds makes it extremely difficult to validate that models are performing accurately at inference time.
This is one of the major barriers to deploying these techniques on mission critical scientific machine learning applications \cite{rudin2019stop}.

This is not to say that there is no work in interpretable or uncertainty-aware scientific machine learning.
In \cite{weinan2020}, it is shown that a 2-layer MLP {\sl can} achieve approximation-theory-like error bounds with a number of parameters that grows with the non-smoothness of $f$, but independently of dimension.
However, the existence of such a model does not guarantee that it will be found during training, or that it even {\sl can} be found without exponential amounts of training data.
However, this result motivates the usage of MLPs for many scientific machine learning tasks, which were previously solved via classical approximation theory, such as those listed in previous sections.

Other works have proposed the usage of {\sl ensemble methods} to provide uncertainty information \cite{maulik2023quantifying}, which could be used for error bounding.
Some analytic error estimation for neural networks also exist, typically in domain specific contexts \cite{de2022generic,guhring2020error,shin2020error}.

\subsubsection{Our Methods for Training Neural Network Models}
\label{sec:nn-methods}

In this work, we are focused on validating MLPs, with the understanding that this could be leveraged for validating portions of more modern architectures.
Given the formulation in (\ref{eq:mlp-model}), there is still significant flexibility in architecture definitions when training MLP models, including the number of layers, weights per layer, and choice of activation function $u(\cdot)$.
There are also significant choices to be made when training.

One of the most common activation functions $u(\cdot)$ for constructing MLPs is the rectified linear unit (ReLU) function, which is what we use in this work.
MLPs with ReLU activation (ReLU MLPs) are also piecewise linear models, which are structurally similar to the Delaunay interpolation models from Section~\ref{sec:delaunay-interp}.
In particular, when trained to (approximately) zero training error, they become piecewise linear interpolants.
In this work, we have trained ReLU MLPs with 3 hidden layers of 100 weights each and ReLU activation functions via \texttt{scikit-learn}~\cite{scikit-learn}.
For training, we use the mean-squared error loss function, a batch size of 20 for backpropagation, and the Adam optimizer with a learning rate of $10^{-8}$.
We also use a randomly selected validation set consisting of 10\% of the provided training data during training, and use early stopping when the validation error drops below $10^{-6}$.
These are recommended settings for \texttt{scikit-learn} and we will see that they work well on a wide variety of problem types in Section~\ref{sec:empirical-study}.
Additionally, to ensure consistency of our results, we have trained 100 models for each problem and selected the best performing weights on the validation set for testing.

While achieving true state-of-the-art performance may require additional work and customization to the problem at hand, we believe that this is a fair and consistent methodology for our studies.

\section{Empirical study of interpolation accuracy}
\label{sec:empirical-study}

In this section, we will consider the problem of interpolating $n$ procedurally generated synthetic data in $\mathbb{R}^d$.
This allows us to consider the effects of the training data spacing and underlying $f$ on interpolation / machine learning accuracy.
We will consider a case study of a real scientific machine learning problem in Section~\ref{sec:sciml-experiments}.
Within this section, Sobol sequence and Latin hypercube sampling (described later) is carried out using the Quasi-Monte Carlo submodule {\tt scipy.stats.qmc} \cite{roy2023}, with the {\tt scramble} option set to {\tt True}.

We have implemented interpolants and error estimates from the previous section in a unified, lightweight Python library~\cite{interpMLrepo}.
As detailed in Section~\ref{sec:interp-overview},
we use routines from {\tt DelaunaySparse}~\cite{changetal2020c} to compute the Delaunay interpolant (\ref{eq:delaunay-interp}) and the revised error bound (\ref{eq:delaunay-bound-rev});
we use {\tt RBFInterpolator} from {\tt scipy.interpolate} \cite{virtanenetal2020} to compute the RBF interpolant~(\ref{eq:rbf-interp-def}); we make a simple approximation to the RBF bound (\ref{eq:rbf-est-gen});
we use {\tt GaussianProcessRegressor} from {\tt sklearn.gaussian\_process}~\cite{scikit-learn} to compute the GP interpolant~(\ref{eq:gp-var}) and associated 95\% confidence interval described in~\ref{sec:gp-bounds}, which, for the sake of simplicity, we refer to as the ``GP bound;'' and
we use {\tt MLPRegressor} from {\tt sklearn.neural\_network}~\cite{scikit-learn}  to train a ReLU MLP regressor as a proxy interpolant.

\subsection{Dataset generation and experimental setup}

Our code allows the creation of random datasets in a deterministic fashion by using seeds to set up a random number generator.
Each dataset we create is uniquely specified by the following parameters:
\begin{itemize}
    \item $d$: dimension of the space from which samples are drawn.
    \item $n$: number of samples to draw.
    \item \texttt{spacing}: Sobol sequence (default), Latin hyper-cube, or uniform.
    \item \texttt{seed}: integer value used to initialize random number generator in numpy
    \item $\omega\geq 0$: parameter controlling the ``variation'' of the response function 
    \item $\alpha\geq 0$: parameter controlling the ``skewness'' of the sampled points
\end{itemize}
Given a set of parameters, the code first draws $n$ samples from $[-1,1]^d$, using the \texttt{spacing} sampling method and a random number generator initiated with the value of \texttt{seed}.
Skew is applied to the data by multiplying the $j$th coordinate of each sample point by $e^{-(j-1) \alpha / (d+1)}$.
By this definition, larger values of $\alpha$ cause greater skew in the data (with more skew in higher indexed coordinates) while $\alpha=0$ causes no skew to the data.

With the points $\{x_i\}$ now set, we compute values of the response function $\{f(x_i)\}$, defined by
\[f(x_i) := f(x_{i1},\ldots,x_{id}) = \frac 12 \left(\frac 1d\sum_{j=1}^d z_j^2 - \prod_{j=1}^d\cos(2\pi\omega z_j) \right),\quad\text{where $z_j := x_{ij} - \frac 12$}.\]
Observe that for $\omega=0$, $f$ is a paraboloid with a minimum value of $-1/2$ at $(1/2,\ldots,1/2)$, which we will refer to as \textit{low variation} over $[-1,1]^d$.
Offsetting the paraboloid from the center of the domain removes one aspect of symmetry and thus makes the function more generic.
As $\omega$ increases, the period of the cosine terms decrease, introducing increased deviation from the paraboloid shape.
When $\omega=1$, $f$ attains two full periods of the cosine terms over each coordinate direction of $[-1,1]^d$, which we refer to as \textit{high variation}.

We have carried out numerous experiments exploring the interpolant errors and bounds over this parameter space and report the most relevant findings below. {We found negligible variations by changing the \texttt{spacing} parameter and hence only report results using Sobol sequence sampling.}

\subsection{Effect of function variation ($\omega$) over $n$ and $d$}
\label{sec:empirical-variation}

In this set of experiments, we look at the performance of the interpolant errors and bounds in four regimes: interpolation or extrapolation when $f$ has either low or high variation.
For ``interpolation'' experiments, we randomly select a point on the $d$-sphere of radius 0.1, which will always lie near the center of the sampling domain  $[-1,1]^d$.
For the ``extrapolation'' experiments, we randomly select a point on the $d$ sphere of radius 2, which is always outside the sampling domain.
We use a Sobol sequence as the sampling strategy and set $\alpha=0$ so there is no skew.
We either fix the dimension to $d=5$ and vary the number of samples, $n$, or fix $n=8192$ and vary $d$; these values were chosen as representative examples for reasonably sized datasets.
We repeat each experiment with five distinct values of \texttt{seed} and report the average of the results.

\begin{figure}[ht!]
    \centering
    \begin{tabular}{cc}
    \includegraphics[width=.48\linewidth]{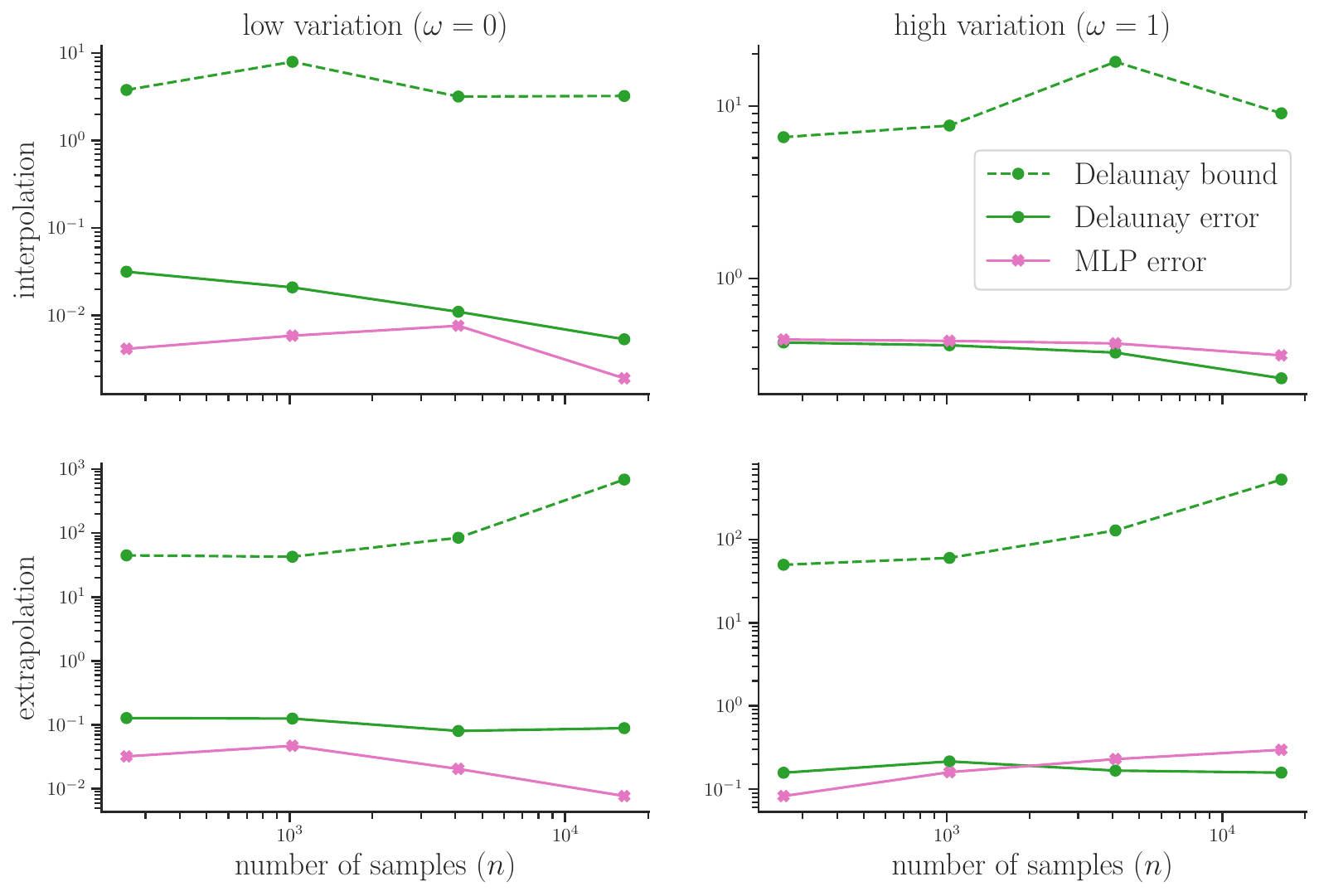} &
    \includegraphics[width=.48\linewidth]{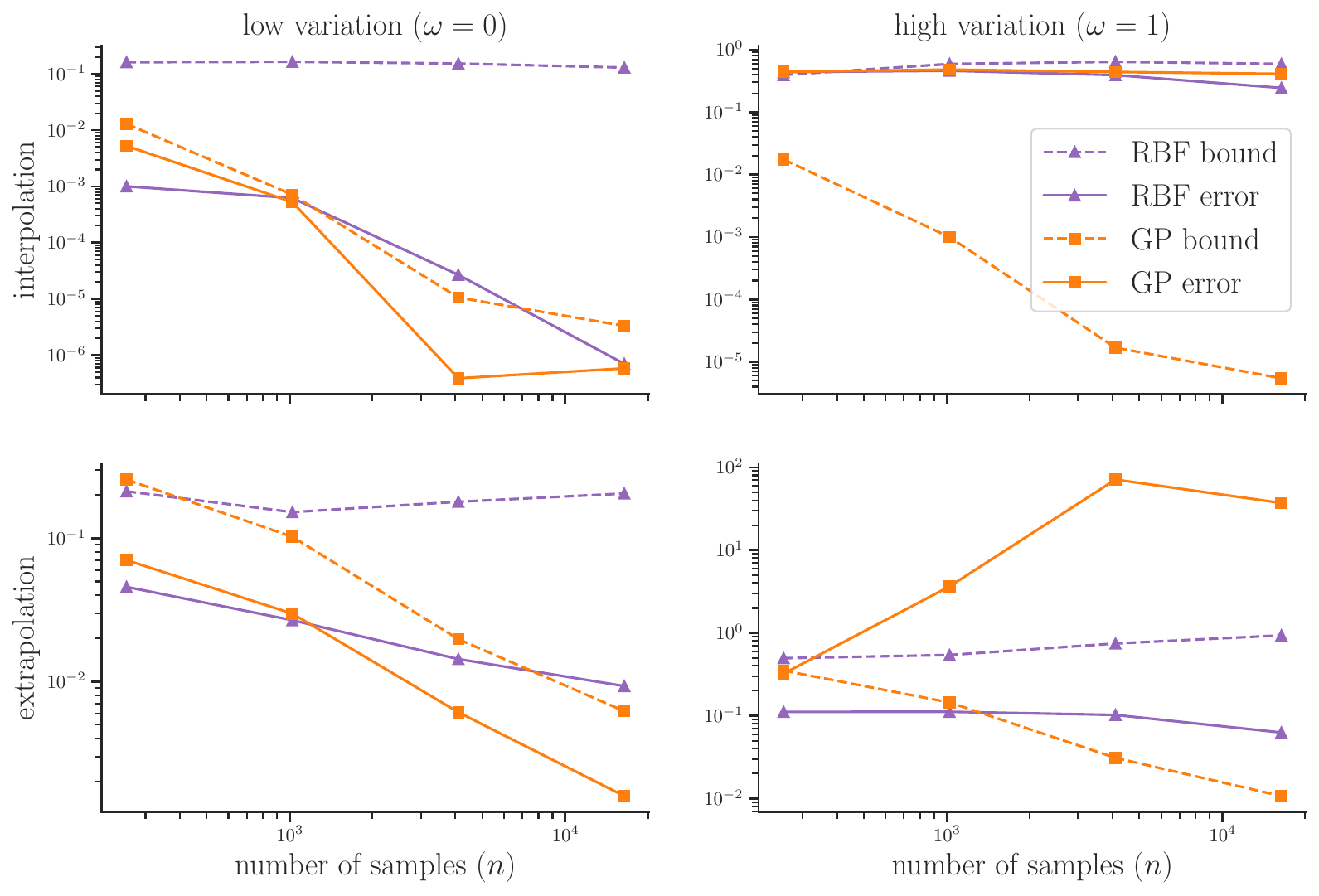} \\
    (a) Delaunay and MLP & (b) RBF and GP
        \end{tabular}
    \caption{For $d=5$ fixed and increasing values of $n$, error bounds and actual error are shown in four regimes: interpolation (top) vs.~extrapolation (bottom) and low variation (left) vs.~high variation (right) in the response function.  Of note, the GP bound is orders of magnitude \textit{below} the actual error in the high variation case.}
    \label{fig:variation}
\end{figure}

In Figure~\ref{fig:variation}, we show the results for fixing $d=5$.
Delaunay and MLP results are shown in (a) while RBF and GP results are show in (b). 
Bounds are shown with dashed lines and actual error (since we know the true function) is shown with solid lines of matching color and markers.
We will keep with these conventions in the remaining subsections.

Turning first to the Delaunay and MLP results in Figure~\ref{fig:variation}a, notice that the Delaunay bound overshoots the actual error by one to two orders of magnitude above in the interpolation regimes and three or more orders of magnitude in the extrapolation regimes.
These experiments suggest that the revised error bound (\ref{eq:delaunay-bound-rev}) may be overly conservative, but that it remains an upper bound in all regimes.
The bound does not improve with the addition of more data, indicating a bound computed from a subsample of a dataset may suffice to bound an interpolant for a full dataset.
Additionally, observe that the MLP approximation performs similarly to the Delaunay interpolant.  
The ReLU MLP is a continuous piecewise linear approximation, just like the Delaunay interpolant, and hence similar behavior between the two is to be expected.

Looking next at the RBF and GP results in Figure~\ref{fig:variation}b, we see markedly different performance between the two interpolants.
The RBF bound is always a strict upper bound for the RBF error, however it ranges from a very sharp estimate in the high variation, interpolation case to an overestimate by five orders of magnitude in the low variation, interpolation case.  
On the other hand, the GP estimate is a relatively tight bound for the actual error in the low variation cases, but dramatically \textit{underestimates} the error, by as much as five orders of magnitude in the high variation interpolation case.  
As mentioned in Section~\ref{sec:gp-bounds}, the poor performance of the GP estimate in this case can be attributed to the fact that $f$ does not satisfy the smoothness assumptions for the estimate.
In an application setting, without \textit{a priori} knowledge of the variation in $f$, the usefulness of the GP estimate is questionable.

Finally, notice the different scales on the vertical axes, especially between (a) and (b).  The RBF and GP errors are three orders of magnitude better than the Delaunay and ReLU MLP errors for the largest $n$ values in the low variation, interpolation case, but this is expected as GP and RBF are providing smooth approximations to a smooth function with ``a lot'' of data.  
The difference in performance between the interpolant types wears off as we move to the extrapolation regime or the high variation regime.
A standout result is the GP error that actually \textit{increases} with $n$ in the high variation, extrapolation case.
This phenomenon is explained by the shrinking support of the basis functions for the GP interpolant as more data is added, which has the effect of removing information from extrapolatory regions.

\begin{figure}[ht!]
    \centering
    \begin{tabular}{cc}
    \includegraphics[width=.48\linewidth]{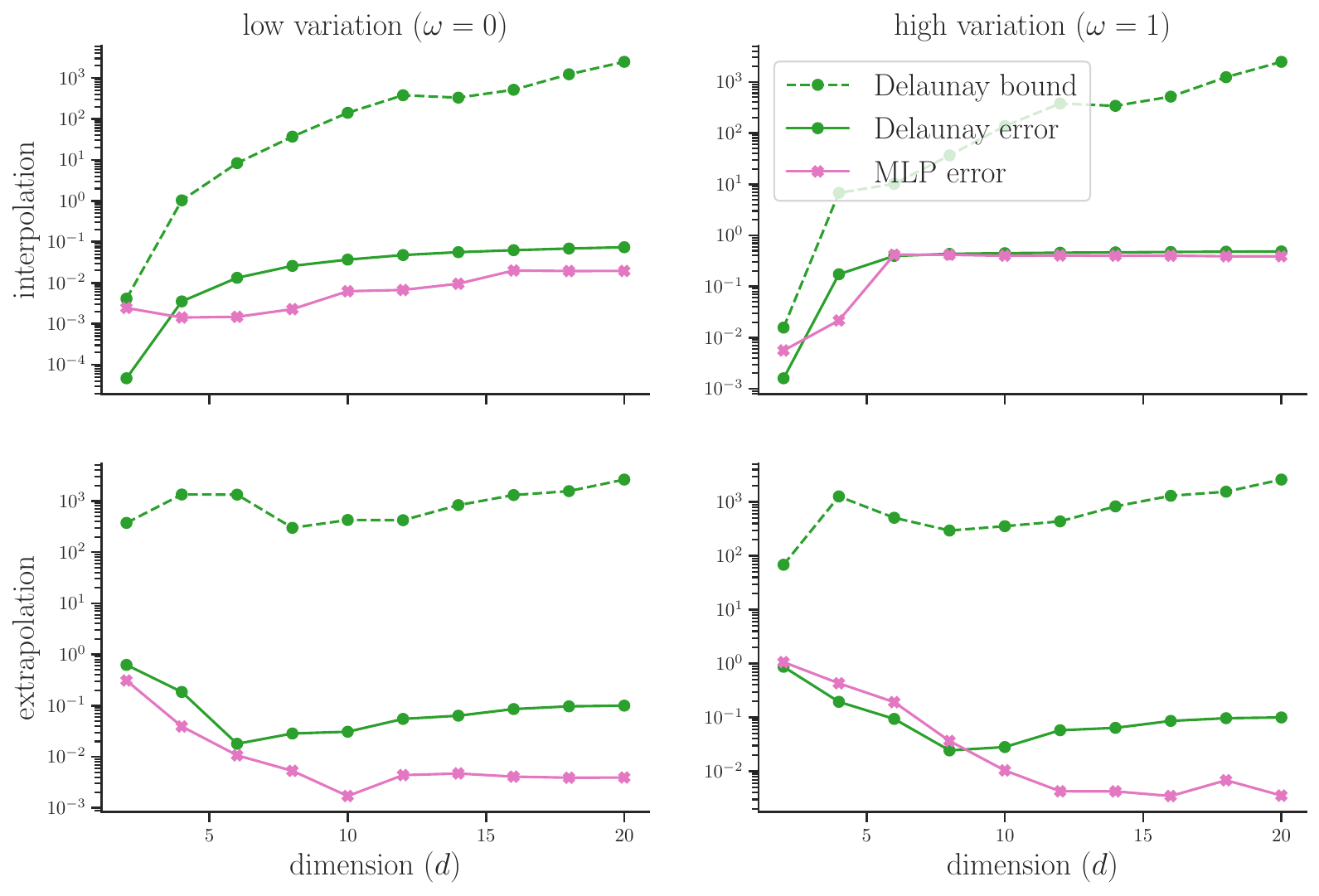} &
    \includegraphics[width=.48\linewidth]{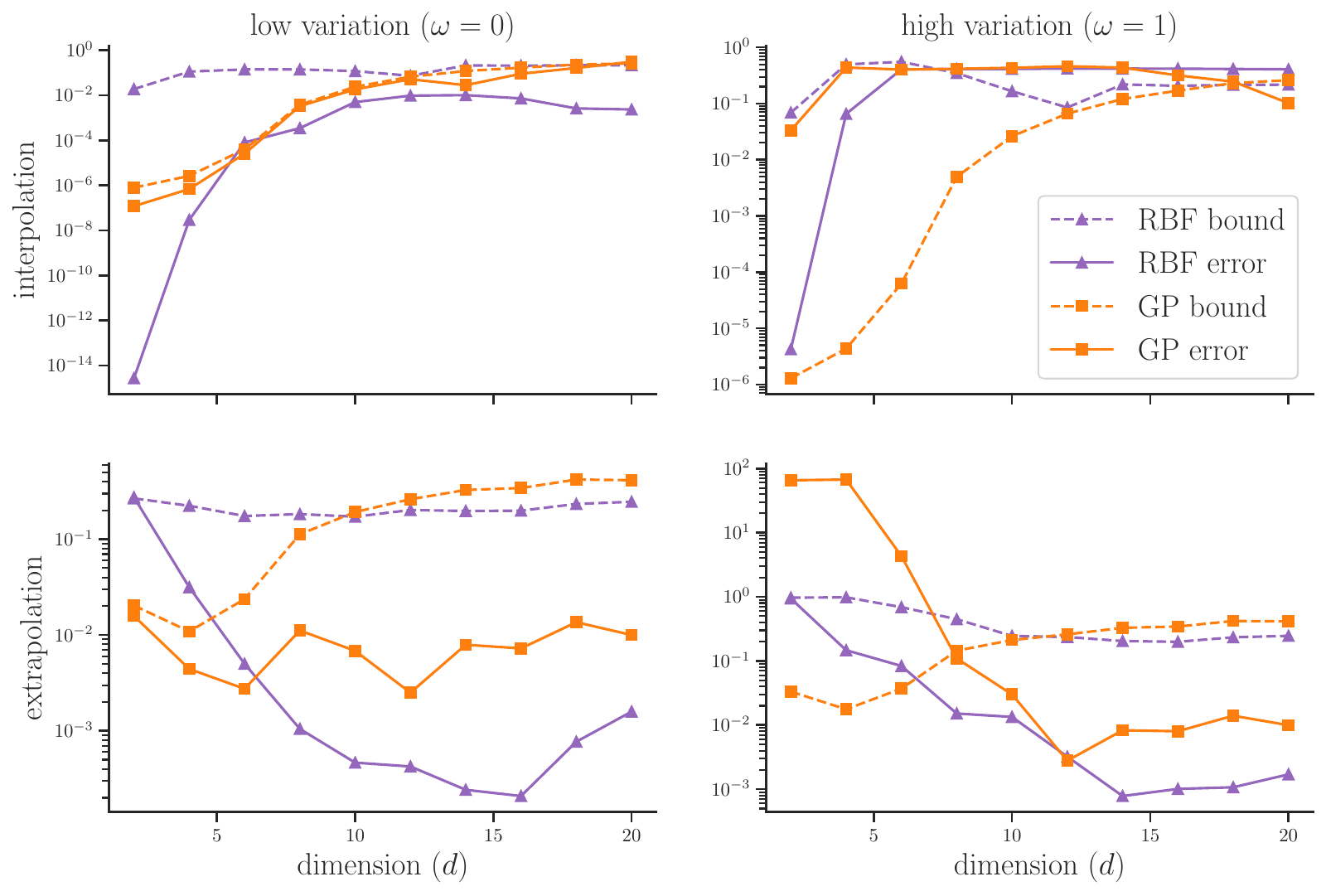} \\
    (a) Delaunay and MLP & (b) RBF and GP
        \end{tabular}
    \caption{For $n=8192$ fixed and increasing values of $d$, error bounds and actual error are shown in four regimes: interpolation (top) vs.~extrapolation (bottom) and low variation (left) vs.~high variation (right) in the response function.  Crossover between the bounds and error as dimension increases is indicative of the exponentially decreasing density for a fixed sample size.}
    \label{fig:dim_vary}
\end{figure}

In Figure~\ref{fig:dim_vary}{a}, we show the results for fixing $n=8192$, with additional conventions as in the previous figure{, for the Delaunay and MLP results. Notice that the relative positions of the error values are mostly stable for $d\geq 5$.}
We infer that the $d=5$ case from Figure~\ref{fig:variation}a is representative of the relative performance of these interpolants for higher dimensions (at least up to 20).  
For $d<5$, we see smaller errors in the interpolation cases, due to the relative density of our fixed $n$ in low dimensions.
As $d$ grows, the error effectively plateaus while the bound increases, reflecting the greater uncertainty inherent in the higher dimensional settings.

For the RBF and GP results in Figure~\ref{fig:dim_vary}b, we notice many orders magnitude lower error of RBFs than any of the other cases in the interpolation regimes, but the benefit is more modest as $d$ increases or as we move to an extrapolation regime.
Also of note, the GP estimate again \textit{underestimates} the error in the high variation case until the dimension gets large ($d\geq 18$ for interpolation, $d\geq 9$ for extrapolation). 
Again, as in Figure~\ref{fig:variation}, the RBF bound is found to be correctly bounding the RBF error, albeit sometimes overshooting by a couple orders of magnitude.

\subsection{Effect of skew ($\alpha$) and the importance of scaling}

In these experiments, we keep the same conventions for interpolation versus extrapolation regimes and vary only the parameter $\alpha$ to adjust the skew in the data set.  
We again use a Sobol sequence, fix $d=5$ and vary over $n$.
We compare data with ``no skew'' ($\alpha=0$) to ``highly skewed'' data ($\alpha=10$) and additionally toggle rescaling of the $\{x_i\}$ points off and on.
Rescaling is commonly used to aid numerical stability (and is \textit{required} for some machine learning algorithms) so we use this opportunity to study its effect in a toy example.

\begin{figure}[ht]
    \centering
    \begin{tabular}{c}
    \includegraphics[width=.8\linewidth]{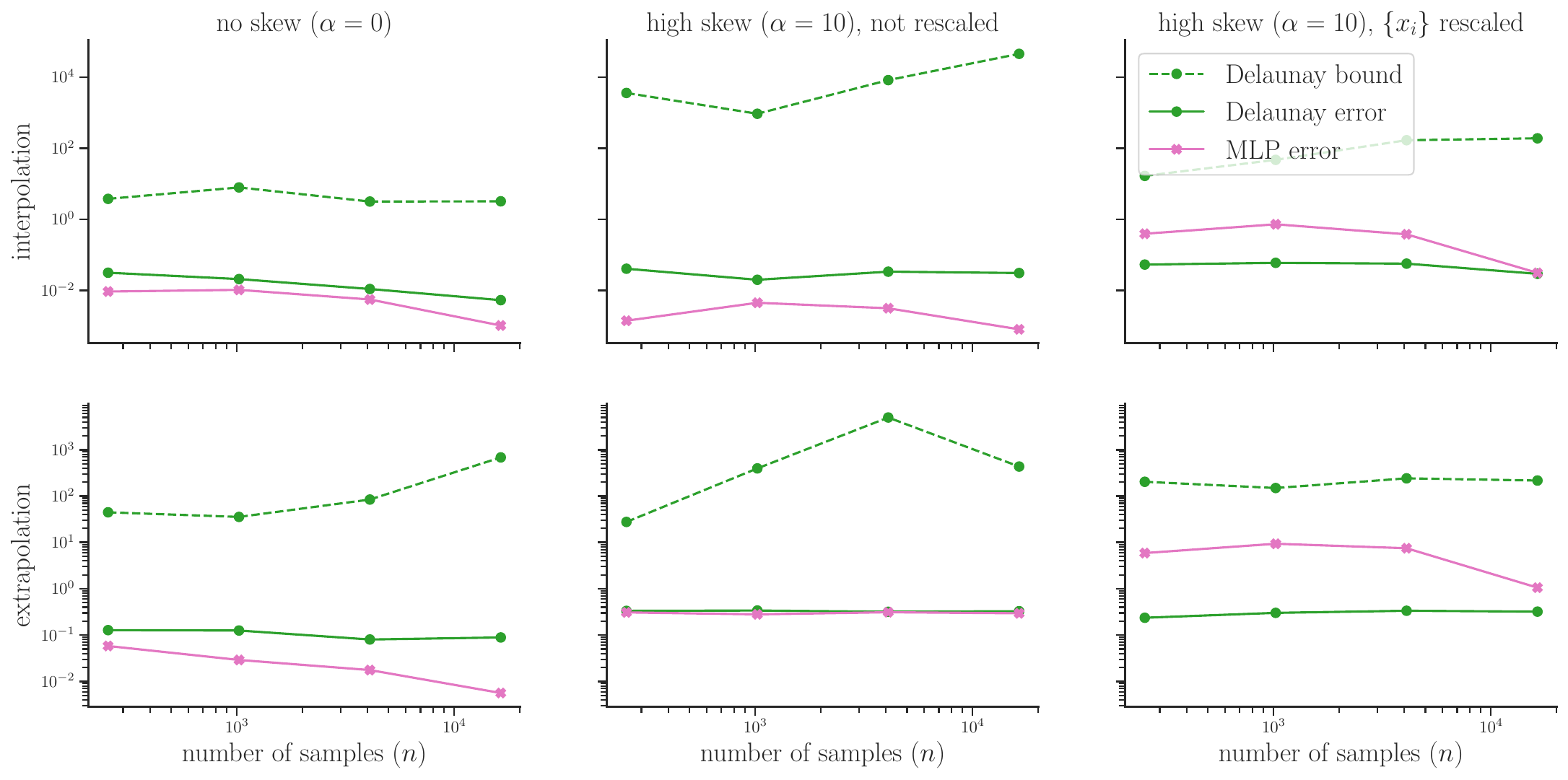} \\
    \end{tabular}
    \caption{Delaunay interpolation is not affected by skew in the input data or rescaling, unlike the ReLU MLP approximant.  The Delaunay bound is affected by skew, but rescaling helps. The left column here is the same as the left column of Figure~\ref{fig:variation}a.}
    \label{fig:skew_vary_del}
\end{figure}

\begin{figure}[ht]
    \centering
    \begin{tabular}{c}
    \includegraphics[width=.8\linewidth]{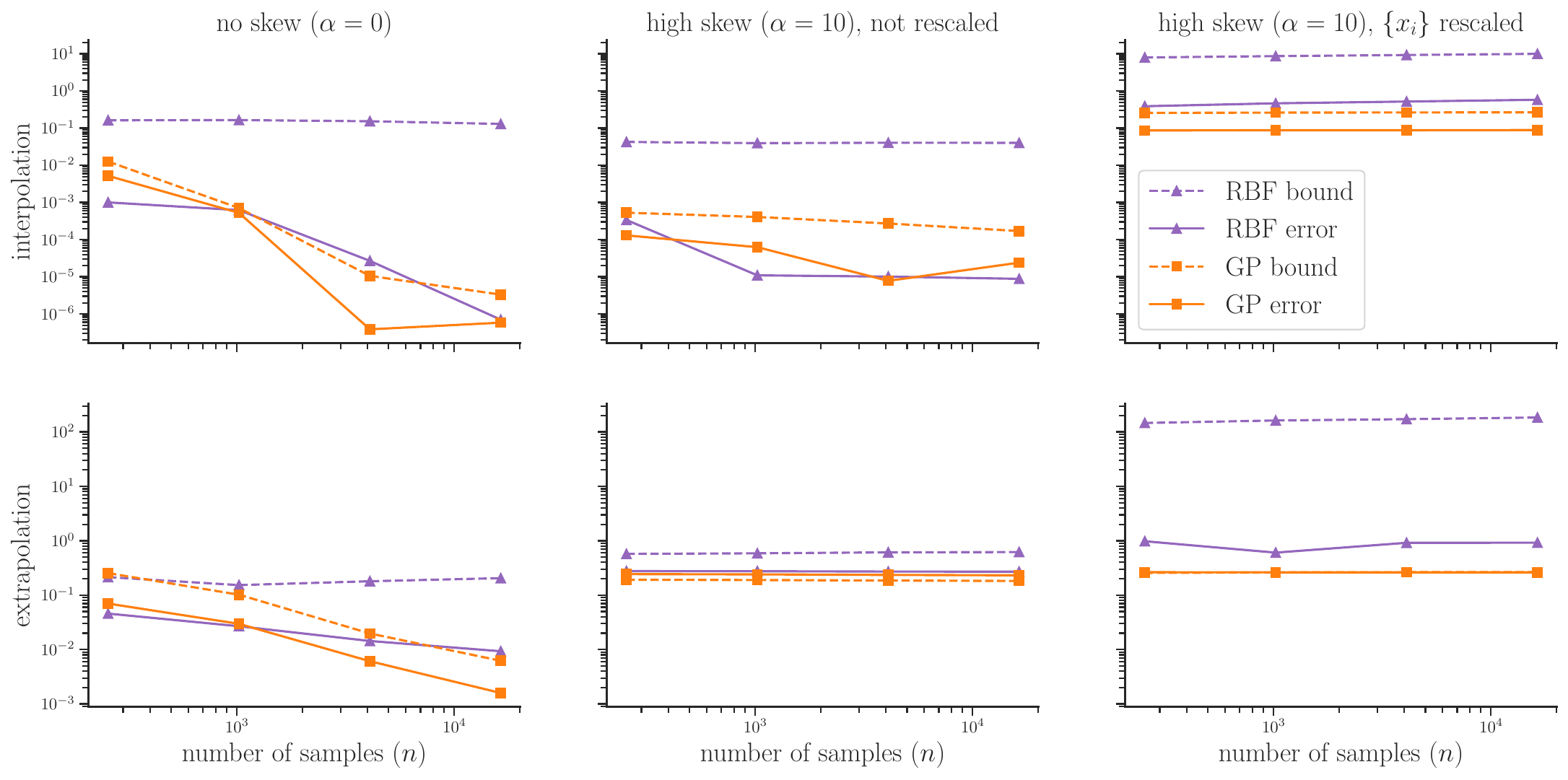} \\
    \end{tabular}
    \caption{GP and RBF interpolants are affected by skew in the input data and rescaling.  If the GP bound does not decrease as the number of samples increases, it may be a sign of skew in the data.  The left column here is the same as the left column of Figure~\ref{fig:variation}b.}
    \label{fig:skew_vary_rbf_gp}
\end{figure}

Figure~\ref{fig:skew_vary_del}, shows the results for Delaunay and MLP.
Note first that the Delaunay \textit{error} is essentially unaffected by the presence of skew, with or without rescaling. 
This behavior is expected as the Delaunay interpolant is defined by geometric computations that are robust against skew and not dependent on scaling. 
The Delaunay bound with rescaling is about an order of magnitude lower than the bound without rescaling (in the interpolation regime), suggesting that rescaling is beneficial for improving the estimate.
Finally, observe that the MLP error is slightly worse in the rescaled case, as MLPs are sensitive to data scaling.

In Figure~\ref{fig:skew_vary_rbf_gp}, we show the performance of the RBF and GP interpolants.
Notice that the GP error and accompanying bound decreases -- by many orders of magnitudes -- when the data is not skewed, but stays mostly flat in the presence of skew, regardless of rescaling.  
Accordingly, if one observes the GP bound decreasing as sample size increases, it can be taken as a sign that the data is (likely) not significantly skewed.
The RBF bound does not capture the improvement in the RBF error from increased data (as was also observed in all cases of Figure~\ref{fig:variation}b) and hence does not provide any signal to the presence of skew.
Finally, note that the rescaling has a dramatic effect on the errors and bounds in the interpolation regime but only significantly affects the RBF estimate in the extrapolation regime.

\subsection{Prevalence of {geometric} extrapolation regimes as $d$ increases}
\label{sec:extrapolation_study}

In our final set of experiments, we explore the prevalence of extrapolation across changes in $n$ and $d$.
Determining the probability that a random test point constitutes extrapolation relative to a fixed data set depends both on properties of the data set and the method by which the test point is selected.
We take a simple approach for both and explain how the takeaway messages are much more than a ``just-so'' story about extrapolation.

\begin{figure}[ht!]
    \centering
    \begin{tabular}{c}
    \includegraphics[width=.9\linewidth]{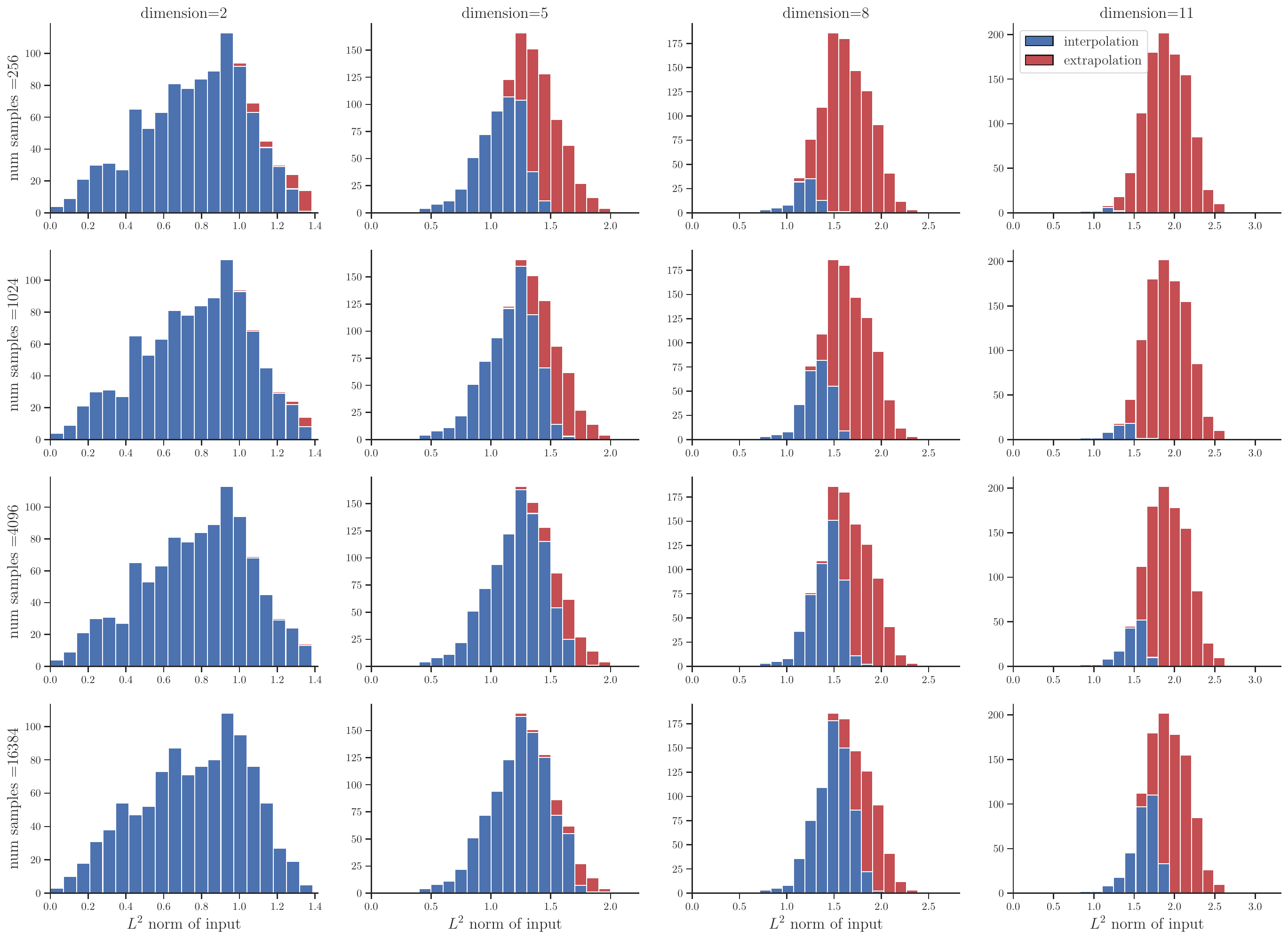} \\
    \end{tabular}
    \caption{Given $n$ (rows) and $d$ (columns), we create a Sobol sequence data set with $n$ points in $[-1,1]^d$.  For 1024 points drawn uniformly from $[-1,1]^d$, we show a histogram of the the $L^2$ norm of each point, colored by whether the point is interpolation (blue) or extrapolation (red) relative to the data set.  Notice that extrapolation always dominates as dimension increases, due to the curse of dimensionality.}
    \label{fig:extrap_hist}
\end{figure}

In Figure~\ref{fig:extrap_hist}, we show the results of the following experiment.  For $n=2^{8+2i}$, $i=0,1,2,3$ (the same values used in Figures~\ref{fig:variation},~\ref{fig:skew_vary_del}, and~\ref{fig:skew_vary_rbf_gp}) we create a data set using a Sobol sequence of $n$ points in $[-1,1]^d$ for $d=2,5,8,11$.  
We set $\alpha=0$ (no skew).
The value of $\omega$ is irrelevant as response values are not used in this study.
For each $d$, we draw $2^{10}$ points uniformly from $[-1,1]^d$, which we call the ``test points.''
For each test point $q$, { we determine whether $q$ lies inside or outside the convex hull of the data set and mark it as an interpolation or extrapolation query, respectively.  We also compute $||q||$.}
Finally, we plot a histogram of the $||q||$ values, using 20 bins from 0 to $\sqrt d$ and color each bar according to the proportion of interpolation (blue) vs extrapolation (red) queries.

Multiple important trends in sampling are evident from the results. 
Looking across each row, it is clear that extrapolation queries rapidly become prevalent among the test points as dimension increases. 
Looking down each column, the proportion of interpolation queries is increased by increasing the number of points in the sample, as is to be expected.
However, it is also clear that eventually---perhaps around $d=8$---attempting to mitigate extrapolation queries becomes computationally impractical.  
The bottom right histogram reflects 16,384 samples in $d=11$, but only $\approx30\%$ of test points are interpolation queries.
This quantity of samples is already large in many scientific machine learning contexts, and acquiring an order of magnitude more data may be infeasible for many real-world applications.  
A more practical approach to mitigating extrapolation is to pursue dimension reduction techniques.

Additionally, observe that as dimension increases, the distribution of $||q||$ values becomes more sharply clustered in the interior of the range of possible values (note~$||q||\in\left[0,\sqrt{d}\right]$ for $q\in[-1,1]^d$).
This phenomenon, known as concentration of volume, is sometimes overlooked in the data science community although it is well understood from the mathematical setting; see, for instance, \cite[Theorem 3.2]{vershynin2015estimation} and the references that follow.
The concentration in $||q||$ values is due in part to the uniform sampling strategy we selected to draw the test points, however, we observed very similar results when using Latin hypercube or Sobol sequence sampling instead.  
Weighted sampling strategies could help distribute $||q||$ values more evenly, however, whether that is beneficial or advisable depends highly on the application context.

{The classical geometric definition of ``extrapolation'' employed above is deterministic and dovetails naturally with the $L_\infty$-type bounds we have analyzed.  In statistical learning, the term ``out-of-distribution'' addresses a similar notion, but a single data point can rarely be definitively declared to lie ``outside'' a distribution.  Accordingly, statistical bounds are typically in averaged $L_2$ norms, which can mask geometric trends in the data.}

\section{Case Study: Predicting Lift-Drag Ratios from Airfoil Images}
\label{sec:sciml-experiments}

In this section, we will demonstrate how the interpolation techniques and bounds proposed in Section~\ref{sec:interp-overview} can be applied to real-world scientific machine learning problems. Specifically, we demonstrate how our proposed function approximation technique may be utilized for surrogate modeling to bypass the solutions of complex high-dimensional partial differential equations.  The data and code used for our case study is available in the Github repository associated with this paper~\cite{interpMLrepo}.

We start with a function approximation task for aerospace engineering applications. Specifically, the task is to predict lift-to-drag ratios for various airfoil shapes, given training data. In classical modeling workflows, it is typical to deploy numerical partial differential equation solvers to predict this quantity by approximating solutions to the Navier-Stokes equations. These solutions are used to compute the integrated quantity of streamwise and normal forces on the airfoil from which the lift-to-drag ratios can be calculated. In this work, we utilize our function approximation technique to bypass the costly overhead of deploying numerical solvers, by using data-driven methods. 

Our training data for this task is obtained from the UIUC Airfoil Coordinates Database~\cite{uiuc96}. This database provides a collection of coordinates for approximately 1,600 airfoils and their corresponding lift-to-drag ratios. These data correspond to a wide range of applications from low Reynolds number airfoils for UAVs and model aircraft to jet transports and wind turbines. Therefore, rapid predictions for the quantity of interest for this dataset corresponds to potential savings in predictive modeling across several different domains.

{
\begin{figure}[ht]
\centering

\begin{tikzpicture}[
    node distance=2cm and 3cm,
    every node/.style={draw, minimum width=2.5cm, minimum height=1cm, align=center},
    arrow/.style={-Stealth, thick}
]

\node (original) {Original Images};
\node (encoded) [right=of original] {Latent Space $\mathbb{R}^d$};
\node (decoded) [right=of encoded] {Decoded Images};

\node (lift_drag) [below=of original] {Lift/Drag Ratio};

\draw[arrow] (original) -- node[above, draw=none, fill=none] {Encoder} (encoded);
\draw[arrow] (encoded) -- node[above, draw=none, fill=none] {Decoder} (decoded);

\draw[arrow] (original) -- node[left, draw=none] {Original data} (lift_drag);
\draw[arrow] (encoded) -- node[right, xshift=10pt, draw=none] {Interpolation methods} (lift_drag);

\end{tikzpicture}

\caption{{The UIUC Airfoil Coordinates Database provides a mapping from a collection of ``Original Images'' to a scalar ``Lift/Drag Ratio.''  We train an autoencoder (top row), then deploy interpolation methods as mappings from the ``Encoded Images'' to ``Lift/Drag Ratio.''  We use this framework to validate the choice of latent space for the autoencoder, assess the characteristics of a hold-out ``test set'' of images, and interpret results from the interpolation methods.}}
\label{fig:flowchart}
\end{figure}
} 

{
Attempting to predict a scalar value directly from image data is a regression rather than a classification task, suggesting an interpolation approach should be used.  However, interpolating over the space of original images---$\mathbb{R}^{128\times128}$---is well-beyond the feasible range of the interpolation methods considered here.
Thus, we use a combined approach, as illustrated in Figure~\ref{fig:flowchart}. 
First we train an autoencoder to embed the original images into a latent space $\mathbb{R}^{d}$ for some $d\in[2,4,6,8]$.  Then, we deploy our interpolation methods to approximate the mapping from the latent space to a lift-drag ratio value, that is, from $\mathbb{R}^d$ to $\mathbb{R}$.
}

{ The autoencoder is implemented as follows.  We utilize two-dimensional convolutional operations alternating with max-pooling operations to reduce the dimensionality of the input image successively via three hidden layers. Subsequently, the two-dimensional (but reduced) representation is flattened and acted on by fully-connected operations to reach the desired dimensionality in the latent space. This procedure is mirrored in the decoder where instead of MaxPooling operations, we utilize UpSampling operations to increase the spatial resolution of the image. In all our hidden layers, both in the convolutional layers as well as the fully-connected layers, we utilize the Swish activation function. The output layer is selected to have a linear activation. All our models are trained for 100 epochs with a learning rate of 0.001 using the ADAM optimizer, with an early stopping criterion that terminates optimization if there is no improvement in validation performance over 10 consecutive epochs. This criterion ensures robustness to overfitting. Additional details can be observed in our associated code~\cite{interpMLrepo}.}

{The data set we use was constructed as follows.  Using the $\sim$1600 airfoil coordinates, 6837 pairings of images and lift/drag ratios were collected (by changing the angle of attack) and shared in a publicly available repository\footnote{https://github.com/ziliHarvey/CNN-for-Airfoil}.  From these parings, a set of 668 data points is held out and designated as the ``test set.''  From the remaining 6169 data points, we set aside 616 points as a ``validation set,'' then train an autoencoder on all or a subset of the 5,553 remaining non-test, non-validation points.}
Note that for all interpolation techniques and the ReLU MLP, we have used the same implementations described in Section~\ref{sec:empirical-study} with input and output rescaling.

\subsection{{ Validation of latent encoding dimension via interpolation methods}}
\label{sec:sciml-validation}

{We now demonstrate how interpolation assessments in the framework of Figure~\ref{fig:flowchart} can help answer two related validation questions when training the autoencoder: ``Was the choice of latent space dimension optimal?'' and ``Was a sufficient amount of training data used?''  In this context, for an image-output pair $(x,y)$, we define:
\begin{align}
\text{true error($x$)}           &:= \left|\left|~\text{interpolation(encoder($x$))}-y~\right|\right|, \\
\text{approximation error($x$)}  &:= \left|\left|~\text{estimated bound(encoder($x$))}-y~\right|\right|, \\
\text{reconstruction error($x$)} &:= \left|\left|~\text{decoder(encoder($x$))}-x~\right|\right|.
\end{align}
For all the autoencoders we trained, the reconstruction error on the validation set was $\approx 10^{-4}$, meaning it provided no information on the quality of the representation learned for different latent dimensions.   
To gain such information, we compute the true error and the approximation error for the various interpolation types discussed.  
We will see that the true error surpasses the approximation error when the latent space dimension is too small and that both types of error inform whether adding more data to the training set improves the embedding.
}

\begin{figure}[ht!]
\centering
\begin{subfigure}{0.55\textwidth}
    \includegraphics[width=\textwidth]{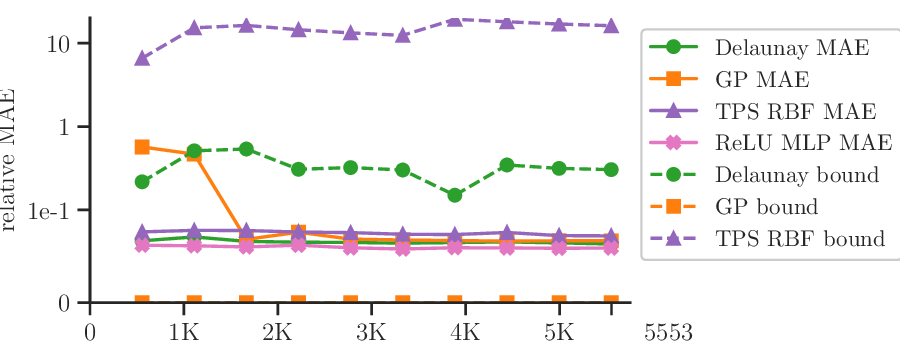} \\
    \includegraphics[width=\textwidth]{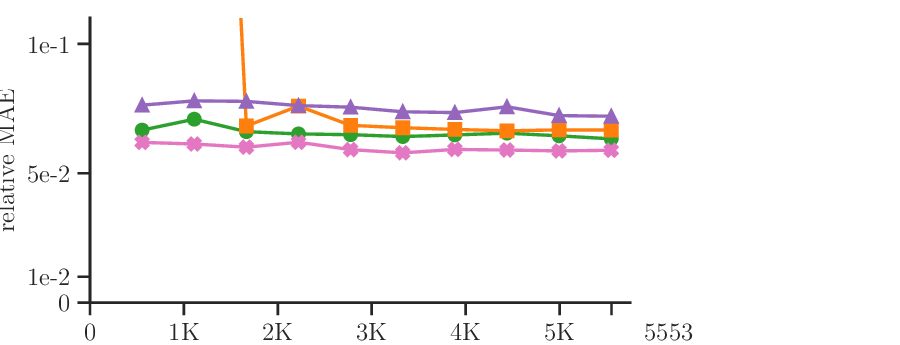}
\end{subfigure}
\begin{subfigure}{0.43\textwidth}
    \includegraphics[width=\textwidth]{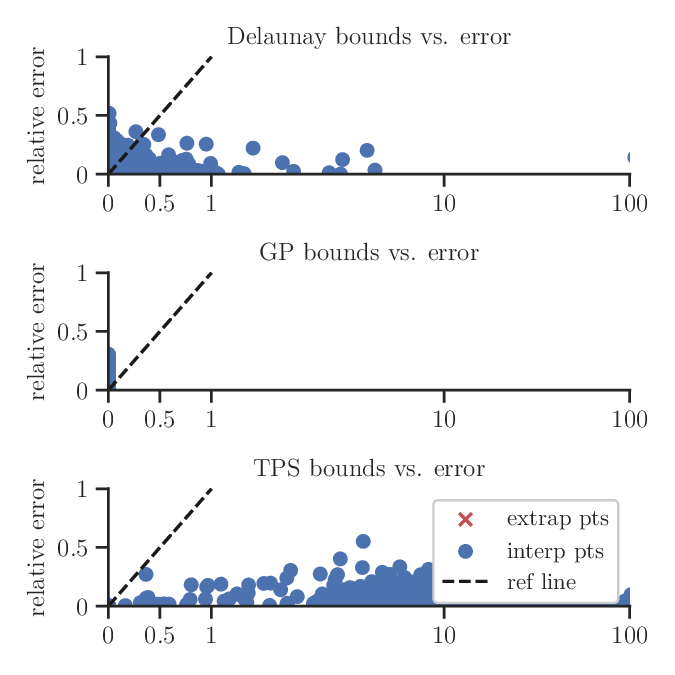}
\end{subfigure}
\caption{Performance with 2-dimensional latent space embedding.
    Top left: MAE and average error bound when predicting lift/drag from geometric shape parameters
    for various training sizes via interpolation and machine learning methods.
    Note that results are shown at semilog scale.
    Bottom left: Only the MAE is shown at linear scale.
    Right: {Approximation error (x-axis) vs.\ true error (y-axis)} for each of the 3 interpolation methods.}
    \label{fig:airfoil2d}
\end{figure}

\begin{figure}[ht!]
\begin{subfigure}{0.55\textwidth}
    \includegraphics[width=\textwidth]{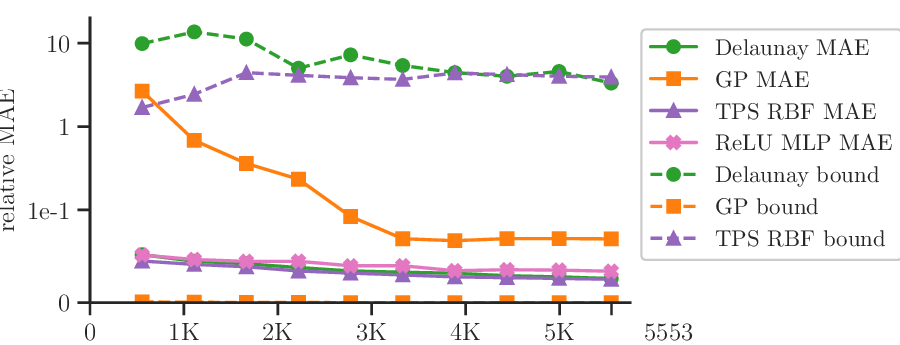}\\
    \includegraphics[width=\textwidth]{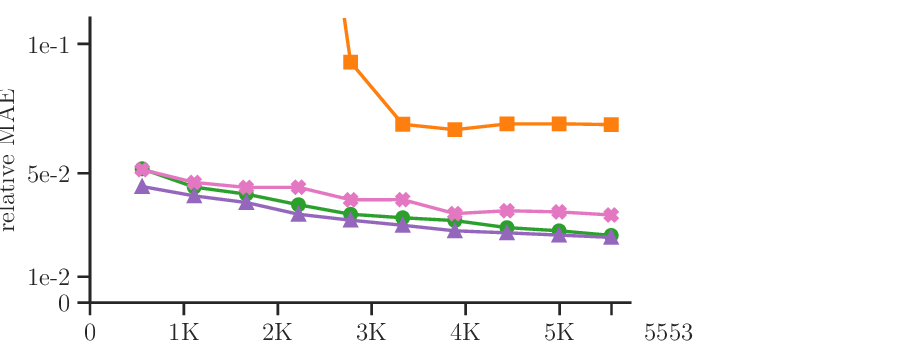}
\end{subfigure}
\begin{subfigure}{0.43\textwidth}
    \includegraphics[width=\textwidth]{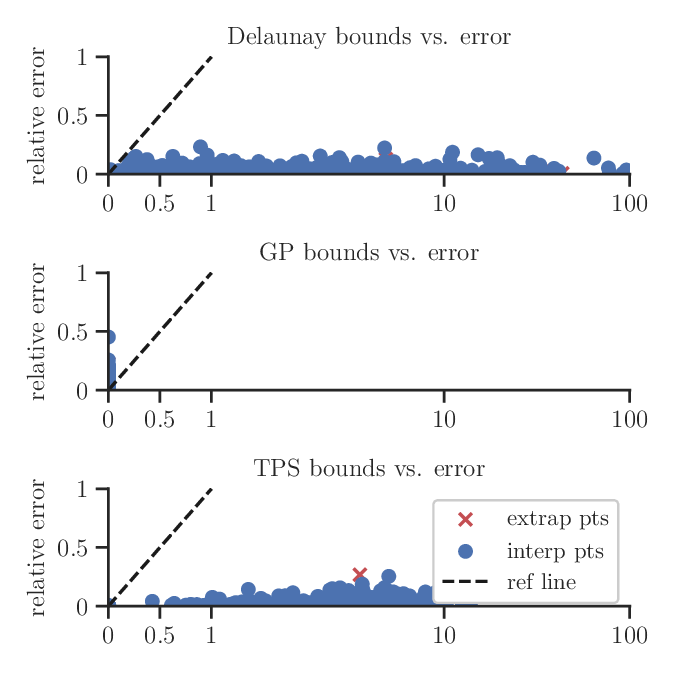}
\end{subfigure}
\caption{Performance with 4-dimensional latent space embedding.
    Top left: MAE and average error bound when predicting lift/drag from geometric shape parameters
    for various training sizes via interpolation and machine learning methods.
    Note that results are shown at semilog scale.
    Bottom left: Only the MAE is shown at linear scale.
    Right: {Approximation error (x-axis) vs.\ true error (y-axis)} for each of the 3 interpolation methods.  See Figure~\ref{fig:lated4D_zoom} for a zoomed in version of the top right plot.}
    \label{fig:airfoil4d}
\end{figure}

\begin{figure}[ht!]
\begin{subfigure}{0.55\textwidth}
    \includegraphics[width=\textwidth]{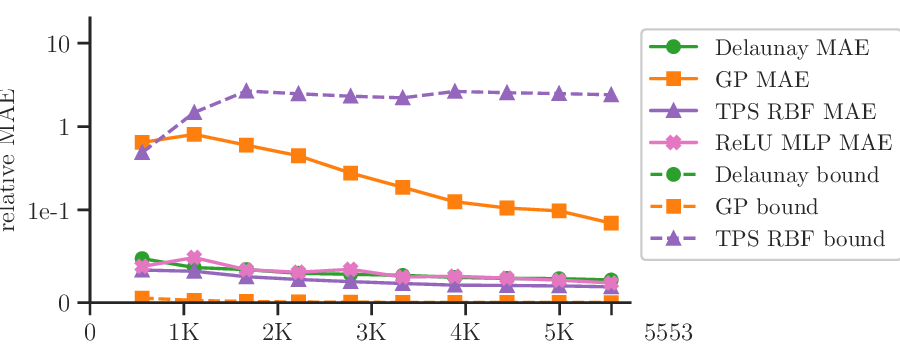}\\
    \includegraphics[width=\textwidth]{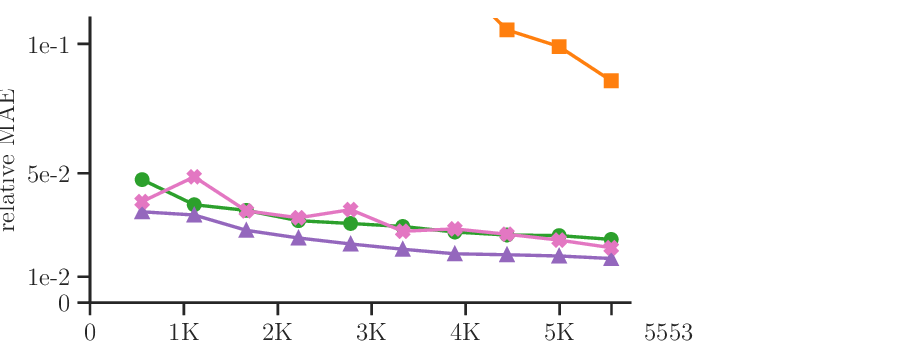}
\end{subfigure}
\begin{subfigure}{0.43\textwidth}
    \includegraphics[width=\textwidth]{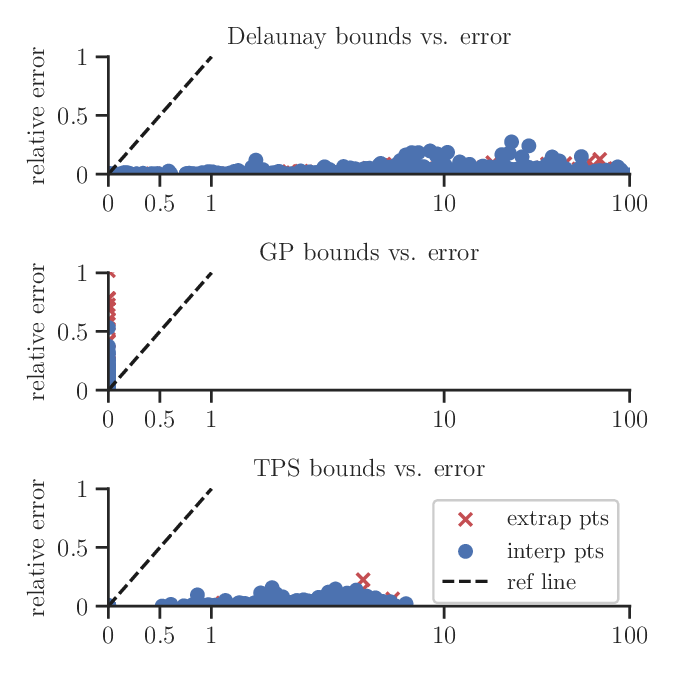}
\end{subfigure}
    \caption{Performance with 6-dimensional latent space embedding.
    Top left: MAE and average error bound when predicting lift/drag from geometric shape parameters
    for various training sizes via interpolation and machine learning methods.
    Note that results are shown at semilog scale.
    Bottom left: Only the MAE is shown at linear scale.
    Right: {Approximation error (x-axis) vs.\ true error (y-axis)} for each of the 3 interpolation methods.}
    \label{fig:airfoil6d}
\end{figure}

\begin{figure}[ht!]
\begin{subfigure}{0.55\textwidth}
    \includegraphics[width=\textwidth]{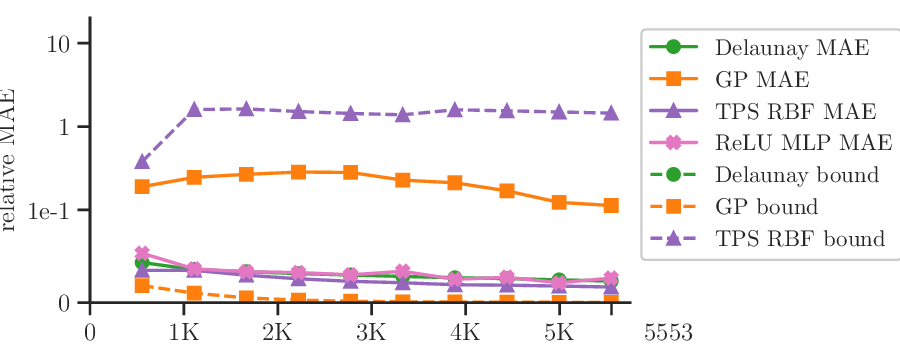}\\
    \includegraphics[width=\textwidth]{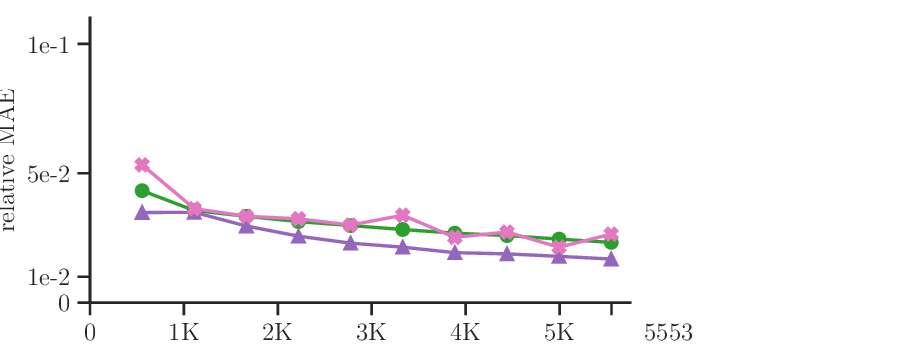}
\end{subfigure}
\begin{subfigure}{0.43\textwidth}
    \includegraphics[width=\textwidth]{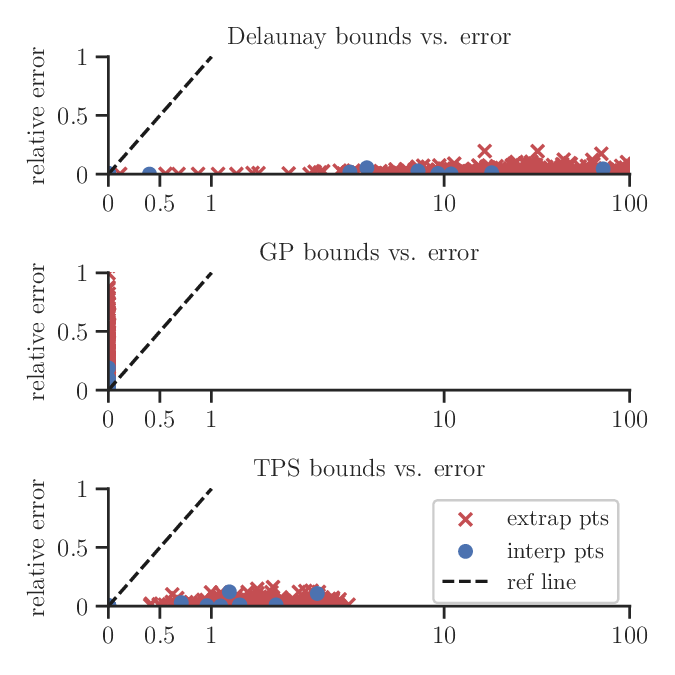}
\end{subfigure}
    \caption{Performance with 8-dimensional latent space embedding.
    Top left: MAE and average error bound when predicting lift/drag from geometric shape parameters
    for various training sizes via interpolation and machine learning methods.
    Note that results are shown at semilog scale.
    Bottom left: Only the MAE is shown at linear scale.
    Right: {Approximation error (x-axis) vs.\ true error (y-axis)} for each of the 3 interpolation methods.}
    \label{fig:airfoil8d}
\end{figure}

{
Our experiments for these validation tasks are as follows.
For each latent space dimension, we train an autoencoder with $p$ percent of the 5,553 training points for $p\in[10,20,\ldots,100]$.  In each trained autoencoder and for each interpolation method and the ReLU MLP, we compute the true error and approximation error at each of the 616 points in the validation set.  We then compute the relative mean absolute error (MAE) of each error type.  We show the results for latent dimension 2, 4, 6, 8 in Figures~\ref{fig:airfoil2d}, \ref{fig:airfoil4d}, \ref{fig:airfoil6d}, \ref{fig:airfoil8d}, respectively.  In each figure, the left plots show MAE values as $p$ increases, at two different scales for the $y$-axis.  The right scatterplots compare the true and approximation errors for the 616 validation points, with extrapolation points marked as X instead of dots.}


One of the first things that we notice is that across all latent space dimensions,
the GP interpolant performs far worse than any of the other methods.
Noting the observations from Section~\ref{sec:empirical-variation},
we comment that this likely indicates that the response surface is nonsmooth in the latent space,
and therefore the Gaussian kernel is not appropriate for this particular problem.
It is likely that we could improve performance by changing the kernel itself for a different option (such as a Matern kernel), but as discussed in Section~\ref{sec:gp-interp}, doing so could interfere with our motivation to use interpolation methods for validation as discussed in Section~\ref{sec:gp-methods}.
Additionally, the GP's 95\% confidence interval regularly shows unreasonably low
expected relative errors.
This indicates that the assumption that response values are Gaussian
correlated with Euclidean distance is likely a poor assumption.

On the other hand, the Delaunay interpolant, TPS RBF, and ReLU MLP all offer
surprisingly similar performance on this problem.
Although the error bounds can be an order of magnitude too high
in many cases (this is expected since they are bounds), they are {always larger than the true error in MAE and almost always larger at individual points.  In particular, we do not observe any points with small approximation error but high true error for these methods.}
Additionally, the Delaunay bounds can be tight when predicting reasonably low error
rates (e.g., relative errors less than 0.5).

From Figure~\ref{fig:airfoil2d}, we see the only case where the Delaunay interpolant's
error bounds fail rather spectacularly.
In fact, in this case they are {\sl anti-correlated} with the true error.
However, we also notice that the true relative error rates are half an order of magnitude lower
for all the other latent space dimensions and continue to improve with added data,
while they do not improve when adding data for this embedding.
This indicates that the $2$-dimensional latent space may not be able to accurately represent
all of the important features in the UIUC airfoil dataset.
We may infer that the encoding is noisy and interpolation methods are not able to provide accuracy below
the noise level.
However, it is worth noting that despite this noise level, the true performance of all methods
remains extremely competitive, indicating that there is little adverse effect from overfitting.

\begin{figure}[ht!]
    \centering
    \begin{tabular}{cc}
    \includegraphics[width=.4\linewidth]{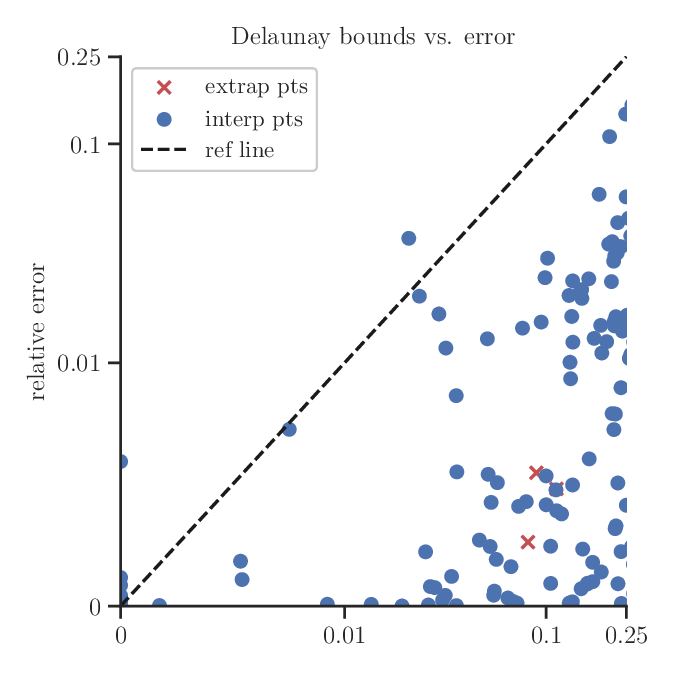} 
    \quad & \quad
    \includegraphics[width=.4\linewidth]{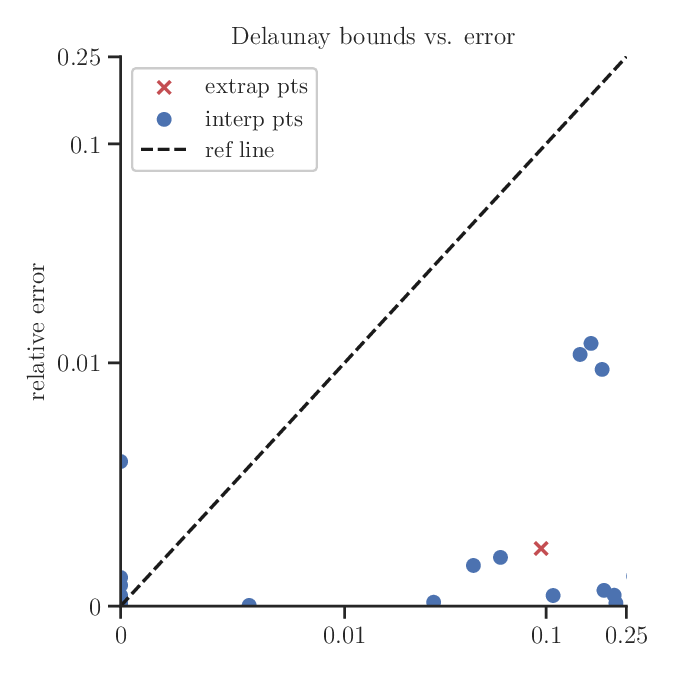} \\
    \end{tabular}
    \caption{{Left: A zoomed in view of the approximation error (x-axis) vs.\ true error (y-axis) for the Delaunay interplant on the 4-dimensional latent space embedding (top right plot from Figure~\ref{fig:airfoil4d}).  The tightness of the bound (and occasional violation) is evident at these small values.  Right: Same figure for the 6-dimensional latent space (cf.~Figure~\ref{fig:airfoil6d}), indicating fewer points with near-zero errors.  }}
    \label{fig:lated4D_zoom}
\end{figure}

For the higher-dimensional embeddings in Figures~\ref{fig:airfoil4d}, \ref{fig:airfoil6d}, and \ref{fig:airfoil8d},
the Delaunay bounds hold, {with few exceptions.  Notably, for the $4$-dimensional case, the bounds appear quite tight near $0$, as seen in Figure~\ref{fig:lated4D_zoom}.
In higher-dimensional latent spaces, these bounds are less tight.}

The bounds for the TPS RBF interpolant remain similar in all dimensions, indicating that the embedded training/validation
points are tightly clustered in the latent space even for higher-dimensional embeddings, since the TPS RBF bound is purely distance based.
On the other hand, the proportion of extrapolation points does increase drastically in high dimensions, which is expected in light of Section~\ref{sec:extrapolation_study}.

In the context of validation, these observations tell us that
\begin{itemize}
    \item the $2$-dimensional representation learned for Figure~\ref{fig:airfoil2d}
is not sufficient for this problem as evidenced by the total failure of the Delaunay error bounds, but the $4$-, $6$-, and $8$-dimensional representations are;
    \item the $4$-dimensional representation is likely to be optimal since it is the lowest-dimensional latent space where
    the bounds appear to hold, indicating that the embedding produces a Lipschitz continuous, low-noise function over the latent space;
    \item when an appropriate latent space is chosen, the practical interpolation bounds given in (\ref{eq:delaunay-bound-rev}) and (\ref{eq:RBF-practical}) are good predictors of errors for certain training points, especially the the Delaunay bound (\ref{eq:delaunay-bound-rev}) for low-dimensional embeddings; and
    \item the interpolation methods---especially Delaunay interpolation---track very closely with the performance of the best ReLU MLP identified via our validation loop, indicating that the interpolation methods provide reasonable expectations for the highest accuracy achievable via a ReLU MLP.
\end{itemize}

Finally, it is worth noting that for these data sets and train/validate splits, the interpolation methods are an order of magnitude cheaper to train and validate than the ReLU MLP.
However, one limitation is that the interpolation methods (especially the Delaunay method) carry a greater proportion amount of their computational cost at inference time, with consequent implications for applications that require real-time predictions.
They also require the training data to be held in memory at all times, which can be a privacy concern.
Therefore, for many applications, { interpolation methods} are best suited as a validation technique.

\subsection{{ Assessing the test set via interpolation methods}}
\label{sec:sciml-testing}

{ We now analyze the 668 points in the ``test set'' using the autoencoder trained on all the 5553 training data with a 4-dimensional latent space.  Looking first at extrapolation prevalence, we find that the geometric distribution of the test data is quite different from the training and validation sets.}
Only 46\% of test points are inside the convex hull of the {encoded} training data, while 98\% of the validation points are inside the convex hull of the {encoded} training data.
{To observe whether this geometric difference can account for the performance of interpolation methods deployed on encoded points from the test set, we report MAE and bound results over the ``in hull'' subset of test points for comparison.}
Results are shown in Table~\ref{tab:af_test_results}.

\begin{table}[ht!]
    \centering
    \begin{tabular}{c|cc|cc}
		 & \multicolumn{2}{c|}{\textbf{All test points}} & \multicolumn{2}{c}{\textbf{In-hull only}} \\
 \bf Method & MAE  & average bound & MAE  & average bound  \\
\hline
Delaunay &  0.21 &  20.04 &  0.20 &  20.77 \\
GP &  17.38 &  0.01 &  0.19 &  0.00 \\
TPS RBF &  0.23 &  12.87 &  0.18 &  6.08 \\
ReLU MLP &  0.21 & n/a &  0.19 & n/a \\
    \end{tabular}
    \caption{Averaged error bounds and MAE for all the {encoded test} data (left) and only {points inside the convex hull of the training data} (right) for all 4 methods from Section~\ref{sec:interp-overview}.  {These results use the autoencoder trained on all the training data with a 4-dimensional latent space.}}
    \label{tab:af_test_results}
\end{table}

One point of note is that the GP performance is actually competitive for interpolation points, but becomes poor when considering extrapolation points.
However, the GP's 95\% confidence interval is still significantly off for all test points.

\begin{figure}[ht!]
\begin{subfigure}{0.49\textwidth}
    \includegraphics[width=\textwidth]{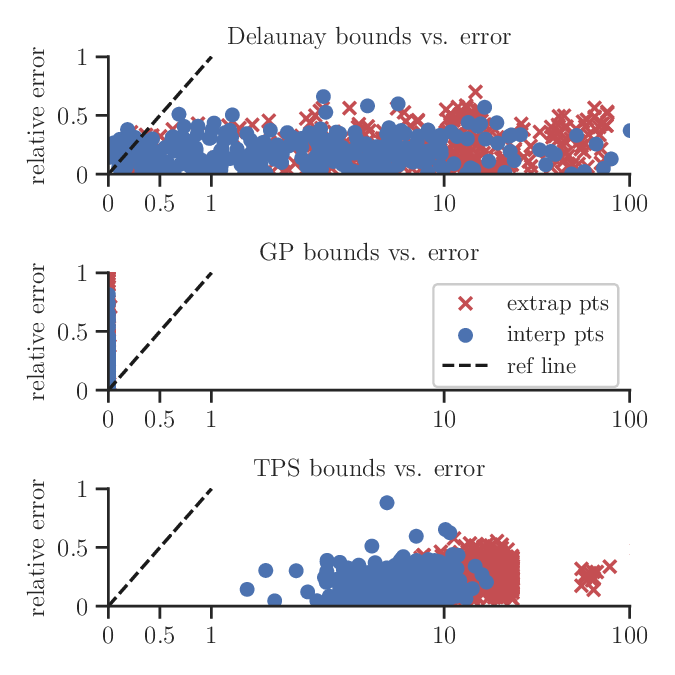}
\end{subfigure}
\begin{subfigure}{0.49\textwidth}
    \includegraphics[width=\textwidth]{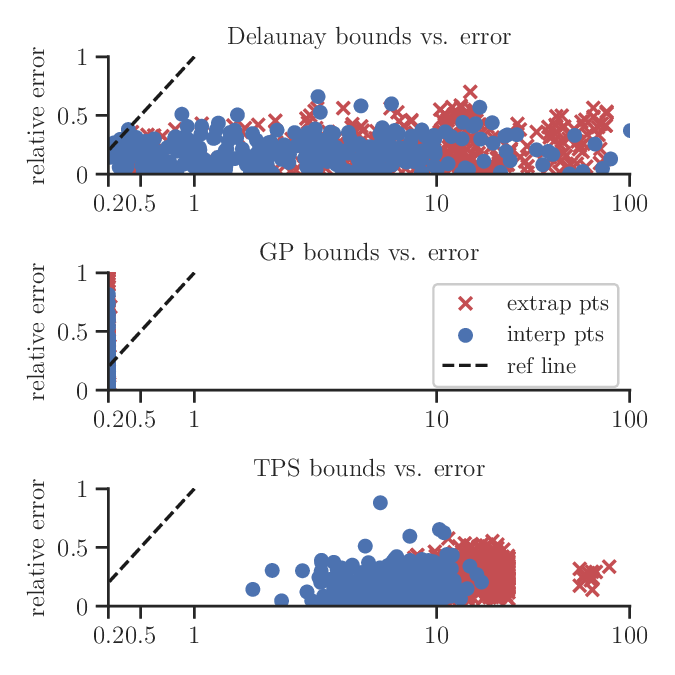}
\end{subfigure}
\caption{Performance and error bounds with the 4-dimensional latent space embedding for 668 testing points.
    Left: Predicted relative error bounds (x-axis) vs.\ observed relative errors (y-axis) for each of the 3 interpolation methods.
    Right: Predicted relative error bounds have been shifted by a fixed amount of $0.2$.}
    \label{fig:airfoil4d_test}
\end{figure}

Compared to errors on the order of $10^{-2}$ obtained when using the full training set during validation in Figure~\ref{fig:airfoil4d}, we see much higher relative testing errors across all categories of Table~\ref{tab:af_test_results} at approximately $0.2$ MAE.
This increase in MAE could be attributed to generalization error and increase in out-of-{hull} predictions.
However, upon inspection of the error bounds, we observe that true errors regularly exceed the error bounds in Figure~\ref{fig:airfoil4d_test}.
Upon closer inspection, the Delaunay bounds are exceeded by approximately $0.2$ for bounds near $0$, which is also the increase in MAE.
Adding a fixed amount of $0.2$ to each relative error bound makes the Delaunay bounds tight again.
This agrees with our hypothesis that the increase is due to generalization error on the test set, which is {largely encoded outside the convex hull of the training data, compared to the validation data}.
We will further explore this in the next section.

Finally, it is worth noting that although not shown here, we also considered testing on the $6$-dimensional and $8$-dimensional latent space encodings from Section~\ref{sec:sciml-validation} to see if they offered improvement.
However, for these embeddings, every test point was far outside the convex hull of the training data in the latent space.
This further demonstrates the distribution shift, and illustrates how quickly the curse of dimensionality takes hold on even moderate-dimensional problems.

\subsection{{ Interpretability enabled by interpolation methods}}
\label{sec:sciml-interpretability}

In this section, we demonstrate the interpretability of the interpolation methods while further investigating the generalization error from Section~\ref{sec:sciml-testing}.
Since the GP was found to be inappropriate for this problem in the previous section, it is not considered here, although a similar approach could be taken were the GP a more appropriate kernel for this problem.
Additionally, one of the limitations of the TPS RBF is that its kernel function cannot be interpreted as a distance metric, making it difficult to leverage here.
Thus, we will primarily focus on the interpretability of the Delaunay interpolant for this problem.

\begin{figure}[ht!]
    \centering
    \includegraphics[width=0.2\textwidth]{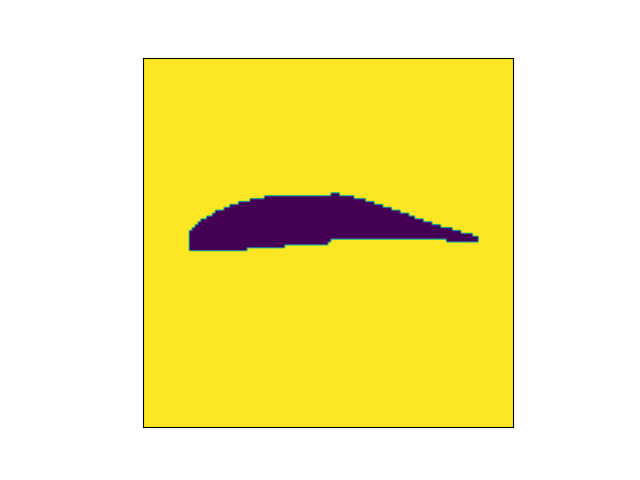}
    $\qquad$
    \includegraphics[width=0.2\textwidth]{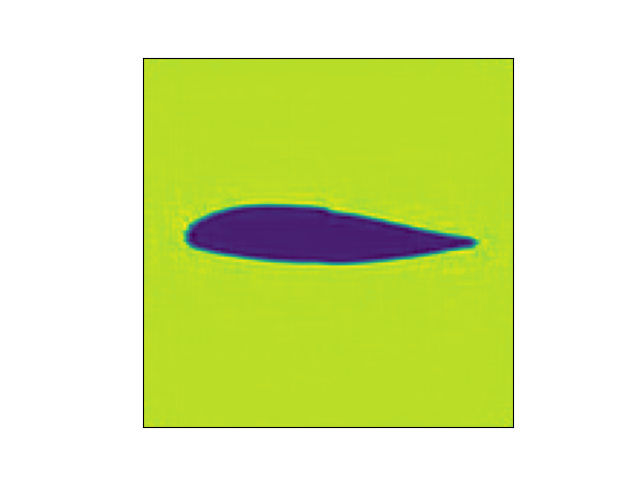}\\

    \includegraphics[width=0.15\textwidth]{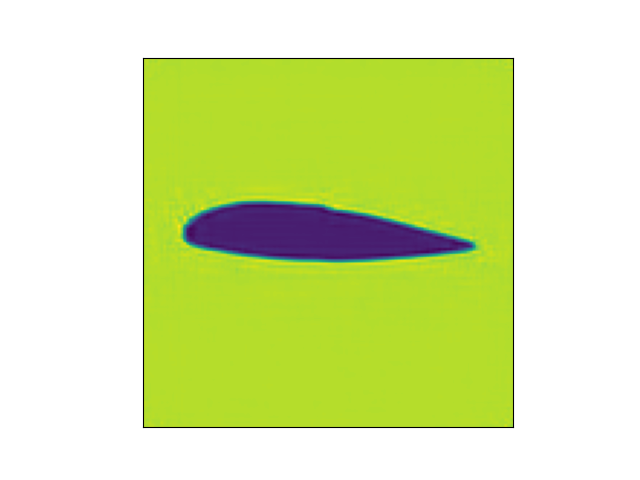}
    \includegraphics[width=0.15\textwidth]{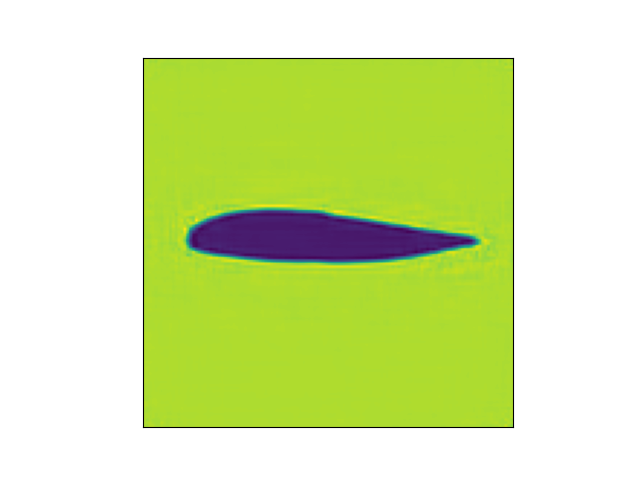}
    \includegraphics[width=0.15\textwidth]{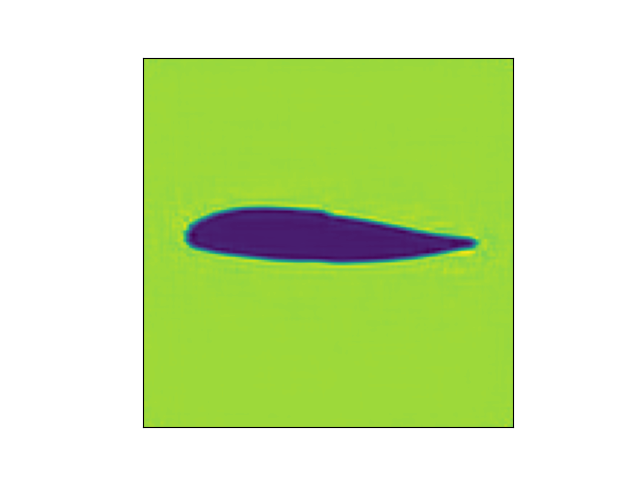}
    \includegraphics[width=0.15\textwidth]{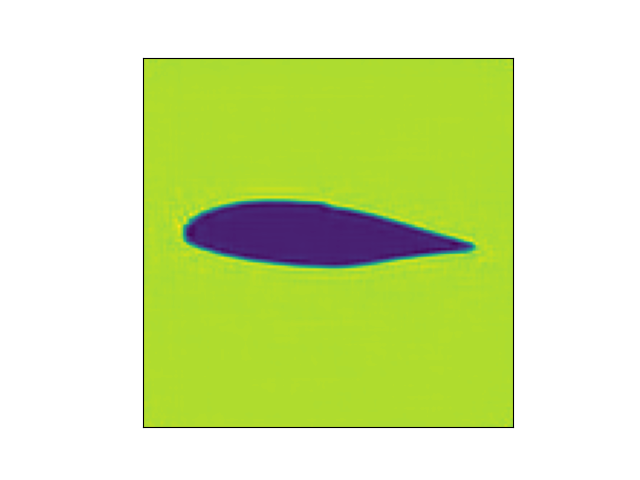}
    \includegraphics[width=0.15\textwidth]{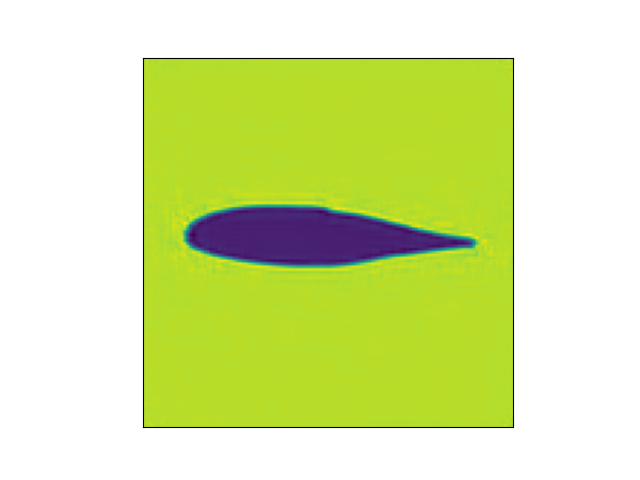}

\caption{Test images where the true error and predicted bounds are both low.
Top: True image (left) and embedding (right).
Bottom: The 5 training images used to make the prediction.
Notice that visually, the embedded image has similar angle-of-attack to the test image and the shape of the top of the foil is also somewhat similar.
However, there is already a clear change in the shape of the foil, signifying an imperfect embedding.
The training images are somewhat similar to the embedding, but not nearly as high-quality as in the validation set.}
    \label{fig:testpt149}
\end{figure}

\begin{figure}[ht!]
    \centering
    \includegraphics[width=0.2\textwidth]{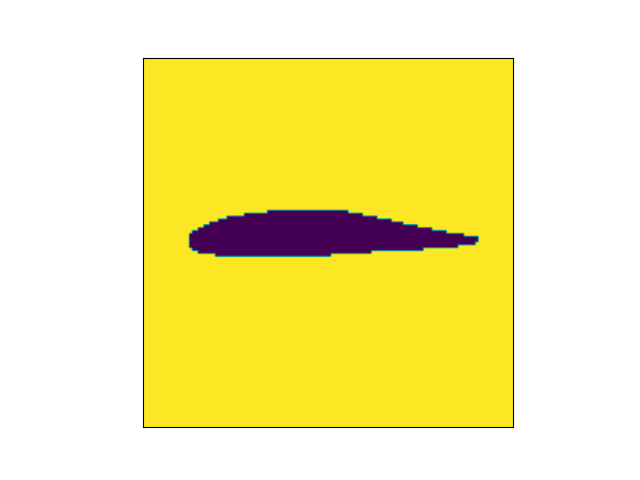}
    $\qquad$
    \includegraphics[width=0.2\textwidth]{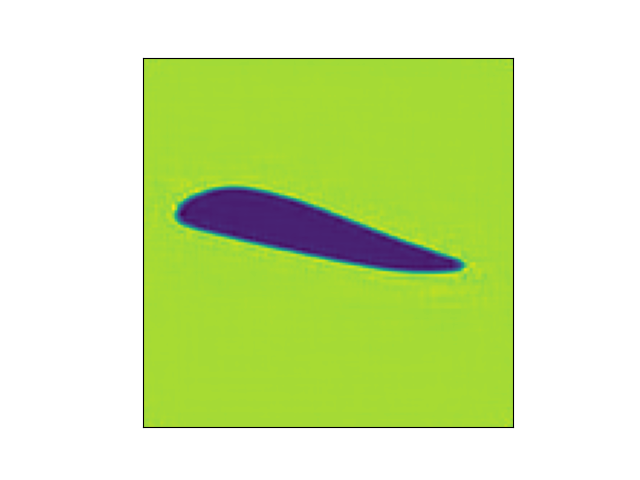}\\

    \includegraphics[width=0.15\textwidth]{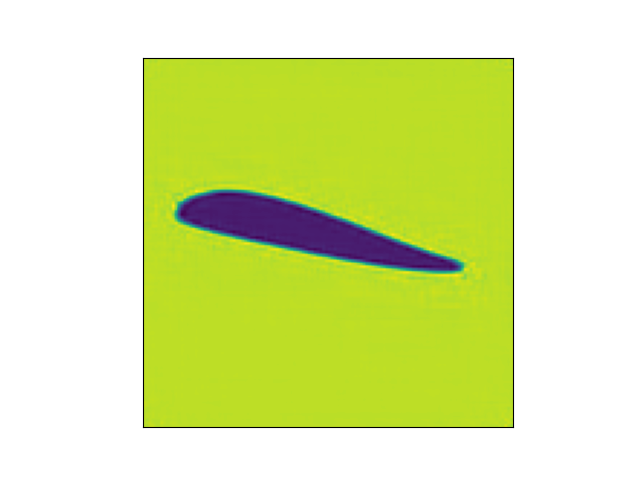}
    \includegraphics[width=0.15\textwidth]{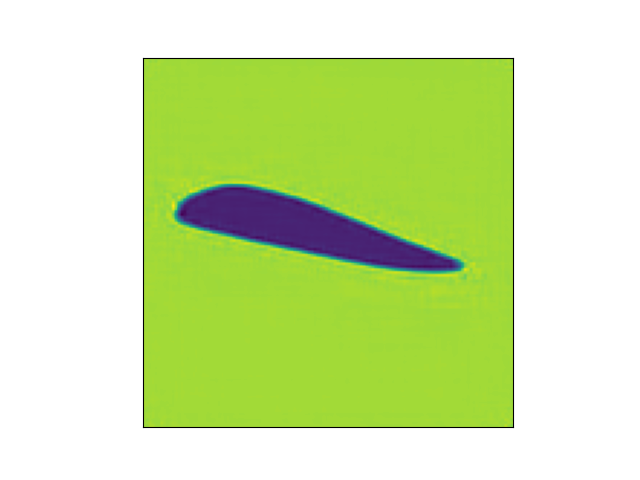}
    \includegraphics[width=0.15\textwidth]{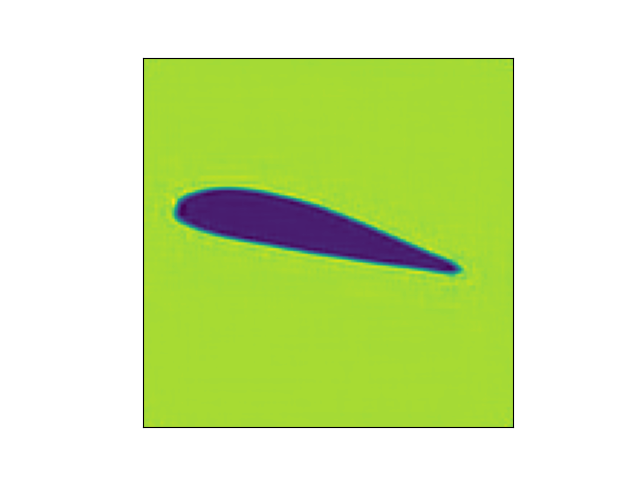}
    \includegraphics[width=0.15\textwidth]{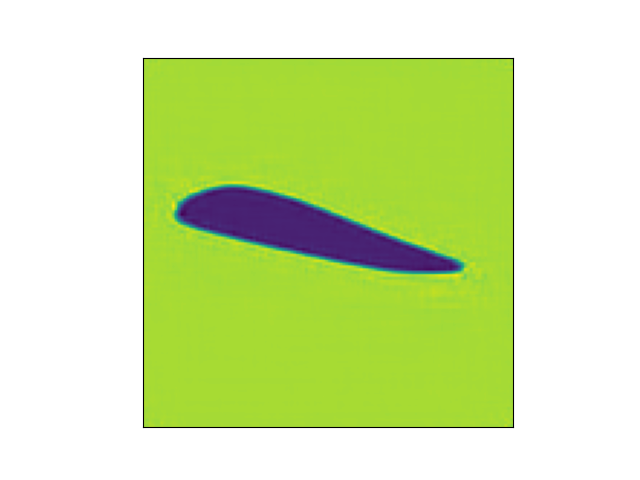}
    \includegraphics[width=0.15\textwidth]{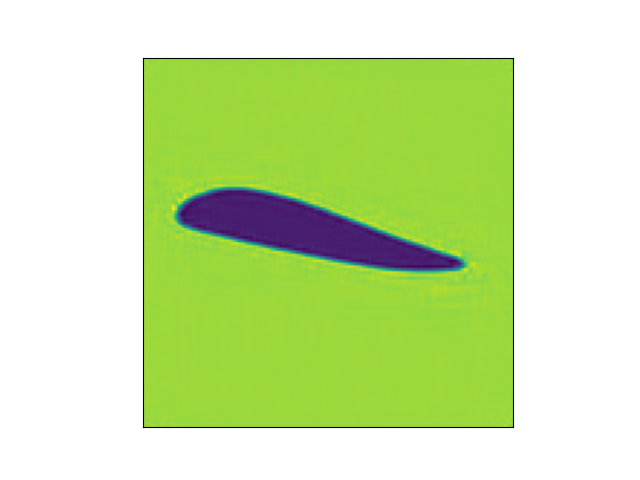}

\caption{Test images where the true error exceeds the error bounds.
Top: True image (left) and embedding (right).
Bottom: The 5 training images used to make the prediction.
Notice that the training images look similar to the embedded image, resulting in a low error bar.
However, the true image is visually very different, resulting in a large actual error.}
    \label{fig:testpt643}
\end{figure}

\begin{figure}[ht!]
    \centering
    \includegraphics[width=0.2\textwidth]{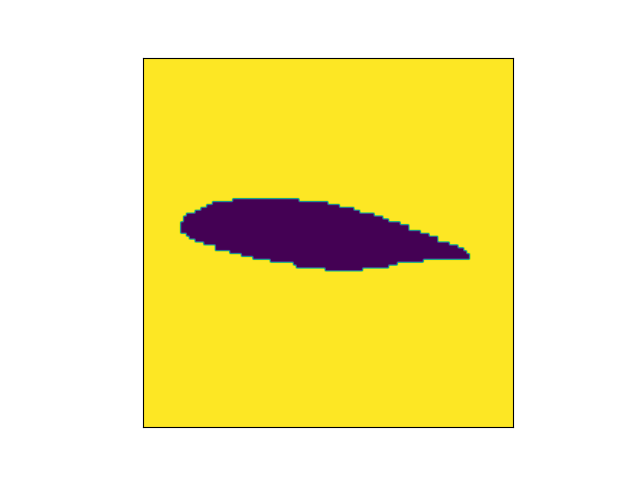}
    $\qquad$
    \includegraphics[width=0.2\textwidth]{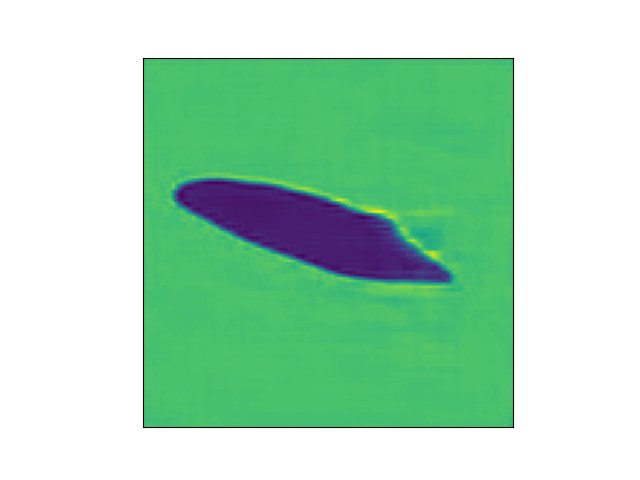}\\

    \includegraphics[width=0.15\textwidth]{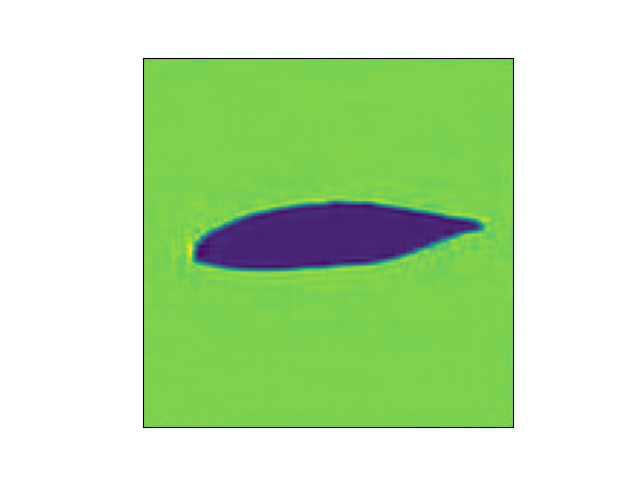}
    \includegraphics[width=0.15\textwidth]{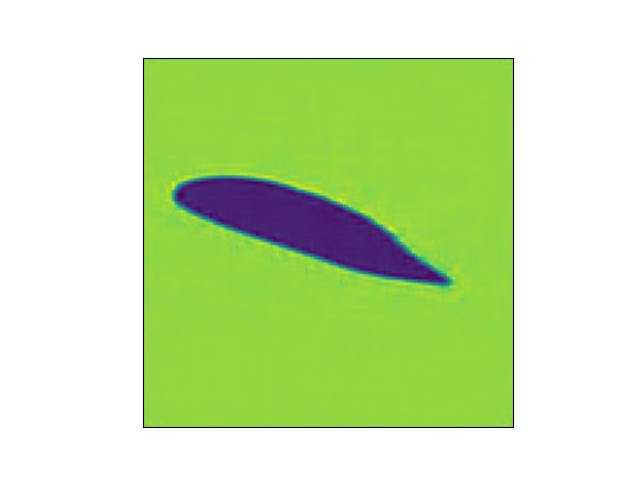}
    \includegraphics[width=0.15\textwidth]{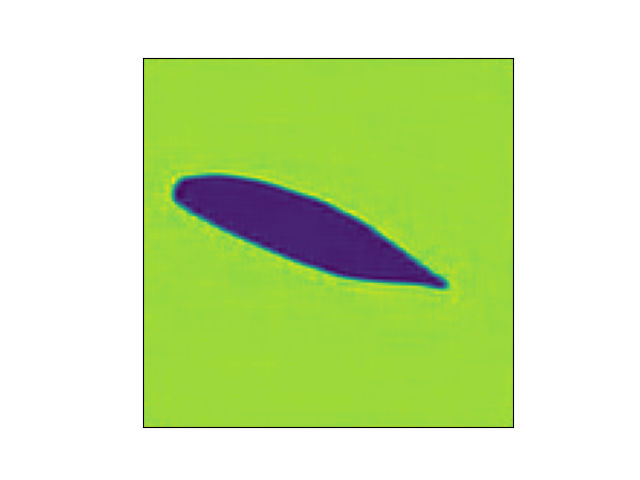}

\caption{Test images where the true error exceeds the error bounds, but the point was flagged as extrapolation.
Top: True image (left) and embedding (right).
Bottom: The 3 training images used to make the prediction.
Notice that the embedding is poor here.
However, the embedded image is also out-of-sample for the training points, resulting in a high error estimate.}
    \label{fig:testpt197}
\end{figure}

For the purpose of this demonstration, three particular points were chosen from the test set to visualize their embedding compared against that of the training points used to make the prediction.
Figure~\ref{fig:testpt149} shows the test point with the lowest prediction error that also satisfies the error bounds.
Figure~\ref{fig:testpt643} shows the test point with the largest prediction error that also violates the error bounds and is an interpolation point.
Figure~\ref{fig:testpt197} shows the test point with the largest prediction error that also violates the error bounds and is an extrapolation point.

428324

\begin{figure}[ht!]
    \centering
    \includegraphics[width=0.2\textwidth]{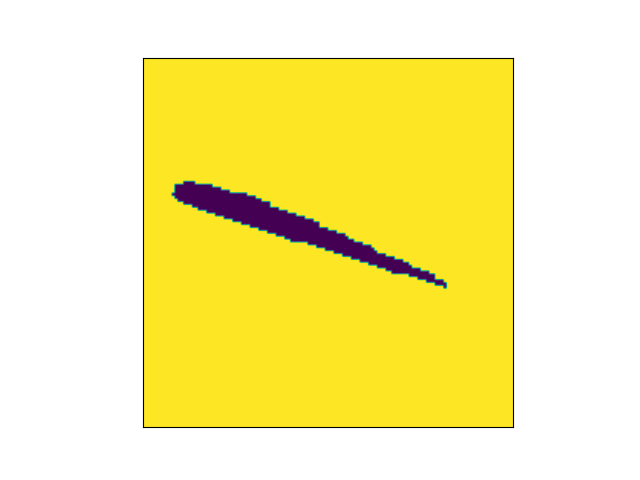}
    $\qquad$
    \includegraphics[width=0.2\textwidth]{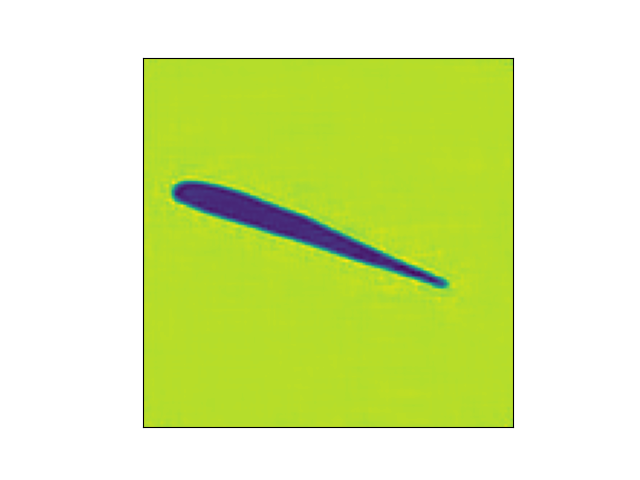}\\

    \includegraphics[width=0.15\textwidth]{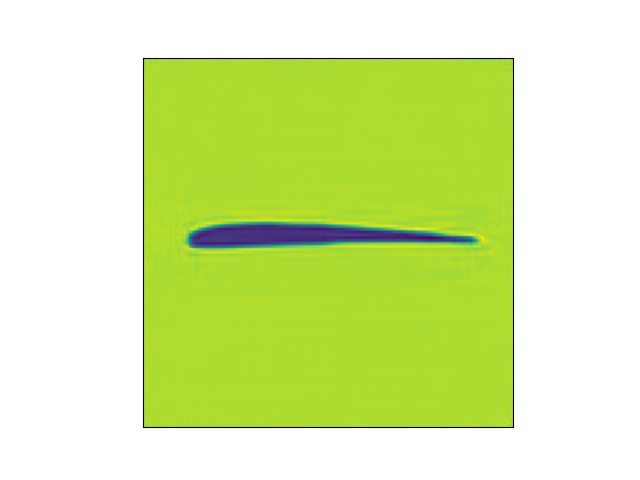}
    \includegraphics[width=0.15\textwidth]{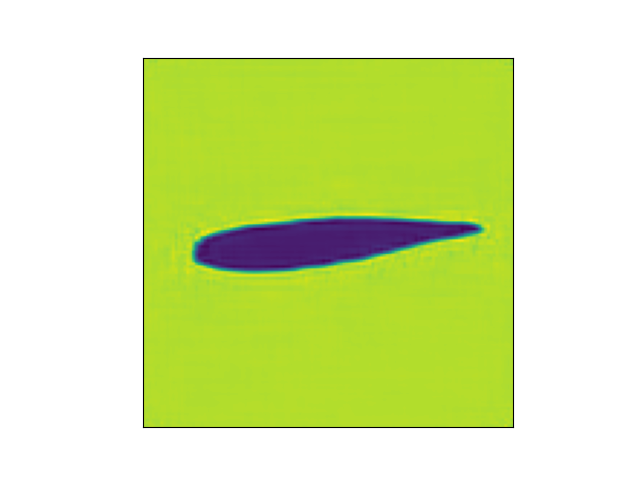}
    \includegraphics[width=0.15\textwidth]{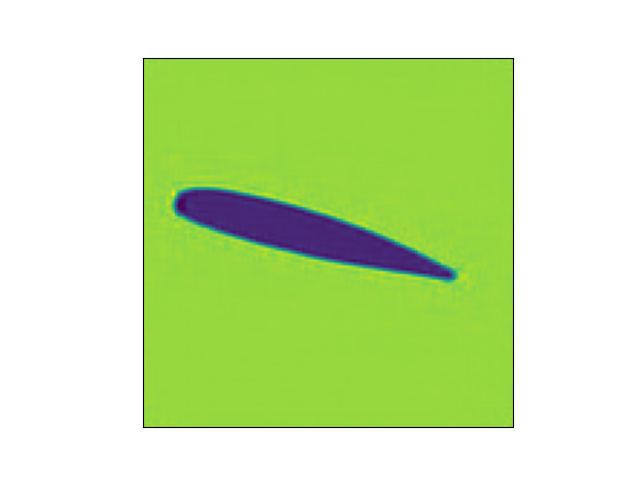}
    \includegraphics[width=0.15\textwidth]{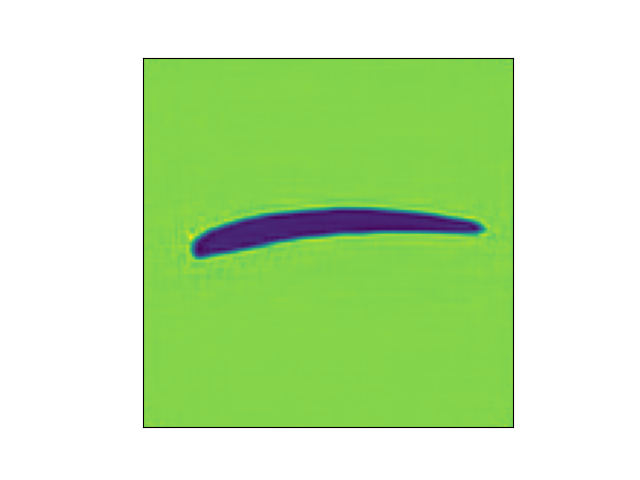}
    \includegraphics[width=0.15\textwidth]{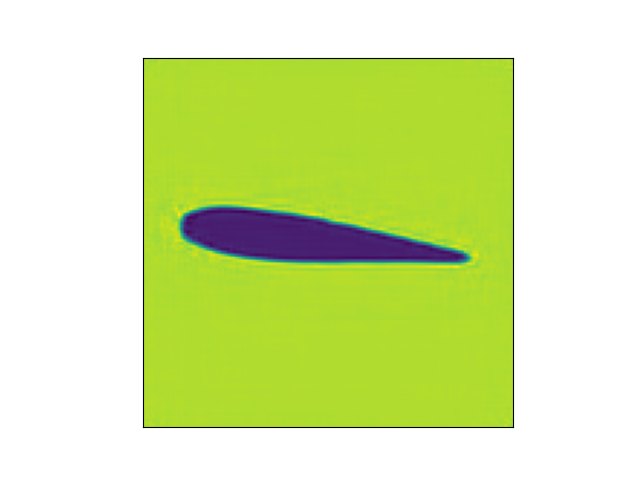}

\caption{Validation image that is also an interpolation point.
Top: True image (left) and embedding (right).
Bottom: The 5 training images used to make the prediction.
Notice that in contrast to Figures~\ref{fig:testpt149} and Figure~\ref{fig:testpt643}, the embedding is very accurate and shares similar shape or angle with several of the training points.}
    \label{fig:valpt605}
\end{figure}

\begin{figure}[ht!]
    \centering
    \includegraphics[width=0.2\textwidth]{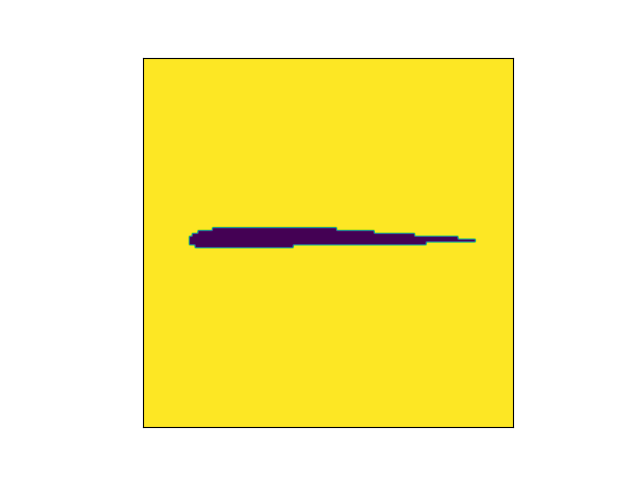}
    $\qquad$
    \includegraphics[width=0.2\textwidth]{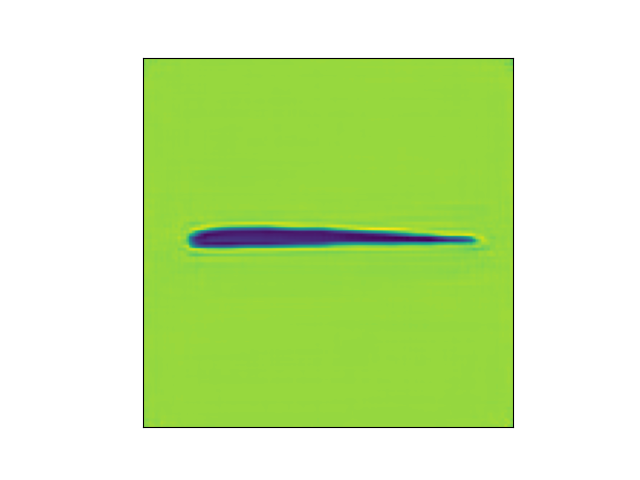}\\

    \includegraphics[width=0.15\textwidth]{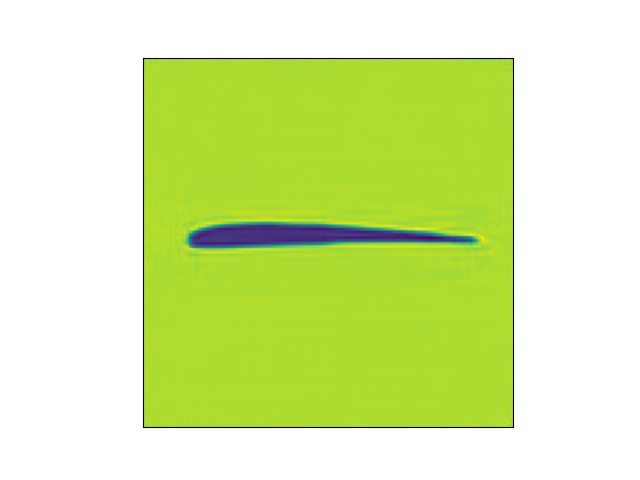}
    \includegraphics[width=0.15\textwidth]{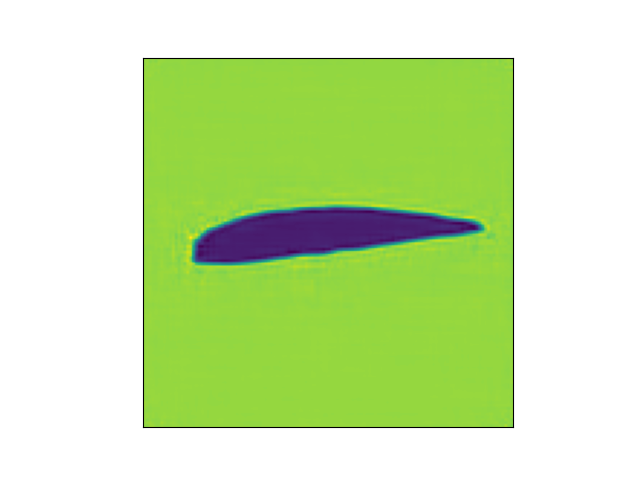}
    \includegraphics[width=0.15\textwidth]{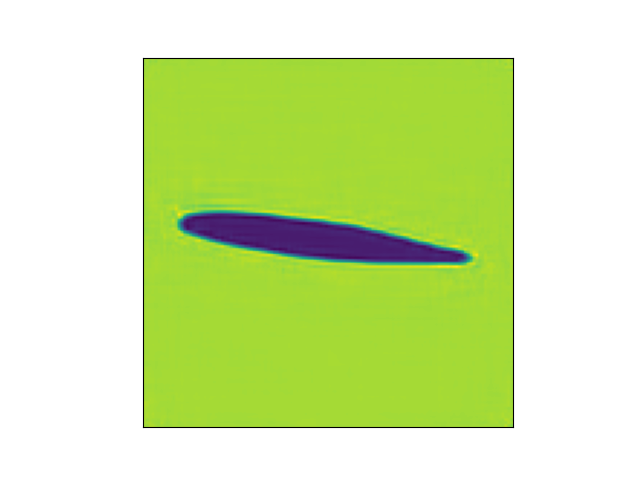}
    \includegraphics[width=0.15\textwidth]{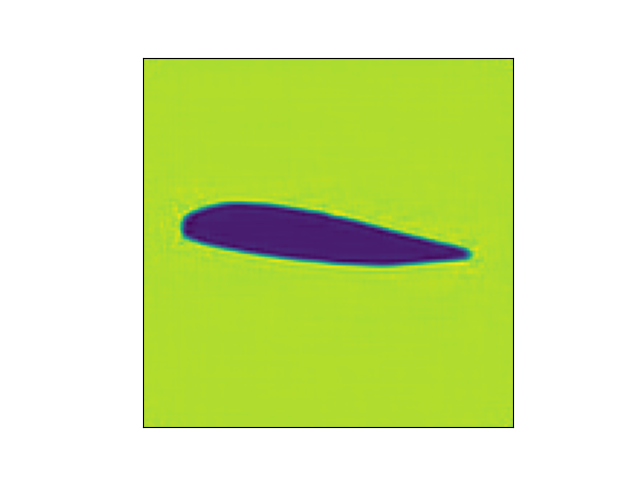}

\caption{Validation images that was was flagged as extrapolation.
Top: True image (left) and embedding (right).
Bottom: The 3 training images used to make the prediction.
Notice that in contrast to Figure~\ref{fig:testpt197}, the embedding is very accurate and very similar to one of the corresponding training points.}
    \label{fig:valpt498}
\end{figure}

First notice, that across all three test points, the embeddings are lower-quality compared to those in the validation set {(see Figures~\ref{fig:valpt605} and \ref{fig:valpt498})}.
This shows that our latent space representation is subject to significant generalization error.
However, the results in Figures~\ref{fig:testpt149}, \ref{fig:testpt643}, and \ref{fig:testpt197} paint a nice illustration of the utility of our methods.
In particular, Figure~\ref{fig:testpt149} (interpolation point that obeys error bounds) shows an embedding that is different but with similar angle-of-attack to the true image, and the training points all closely resemble the embedding.
In Figure~\ref{fig:testpt643} (interpolation point that violates the error bounds), we see an embedding that has almost no similarity with the true image, but closely matches many training points.
In this case, the error in the embedding essentially ``tricks'' the interpolation bounds.
Finally, in Figure~\ref{fig:testpt197}, the true test image is somewhat similar in shape, but not identical and with different angle-of-attack to its embedding.
The embedded image also features an odd distortion on the top right side of the airfoil, which is not present in any of the training images.
Thus, it is visually apparent why this point resulted in geometric extrapolation in the latent space.

In all of the above cases, the interpolation methods would not able to account for a poor autoencoder embedding.
This likely explains why the generalization error was so great, in particular, exceeding the interpolation error bounds even for the Delaunay method, whereas no predictions in the validation set exceeded the interpolation error bounds for the Delaunay method.
In support of the above claim, consider the sample interpolation point and sample extrapolation point from the validation set, shown in Figures~\ref{fig:valpt605} and \ref{fig:valpt498}, respectively.
Note that there is no mention of which points exceed the error bounds here, since on the validation set, every prediction was within the error bounds.

The most striking thing that is immediately obvious from Figures~\ref{fig:valpt605} and \ref{fig:valpt498} is that the embedded images are visually much more similar to the true images than in any of Figures~\ref{fig:testpt149}, \ref{fig:testpt643}, or \ref{fig:testpt197}.
This seems to confirm that it is the failure of our autoencoder embeddings to generalize to the test set, which caused the interpolation error bounds to fail.
In this sense, the failure of the interpolation bounds serves is a feature not a bug, as it serves as an alarm bell for an issue in an earlier stage of our scientific machine learning pipeline.

\subsection{Summary of case study}

In this section, we studied the utility of interpolation methods on a computer-vision-based regression problem from computational physics.
We combined common deep learning techniques (representation learning) with interpolation by training a convolutional autoencoder.
In this particular example, both the TPS RBF and Delaunay interpolant offered similar performance to the ReLU MLP in both validation and testing.
The failure of the GP method indicates that the Gaussian kernel and possibly other assumptions of GP models are not well-suited to this application, most likely due to the smoothness of the embedding.
{An advantage to using these interpolation techniques is their ability to validate our trained ReLU MLP networks against other methods without any training, giving confidence in the predictions of the neural network.}

Furthermore, we made theory-backed choices for the latent space dimension during validation, and quantified the expected errors on the validation set.
On the testing set, we observed significant generalization error {and computed that more than half the encoded testing data lay outside the convex hull of the encoded training data.
We identified that most of the MAE for the true error with the Delaunay interpolant was due to points with near-zero approximation error, and found that adding a fixed amount (0.2) to the Delaunay bound recovered tight bounds and reduced the MAE.}
Finally, leveraging the interpretability of the Delaunay interpolant, we were also able to see that the quality of the latent space embeddings had greatly decreased on the testing set, which is likely a result of this distribution shift.

{ Applying our techniques to a problem with latent space dimension $\mathcal{O}\left(10^1\right)$ is feasible, provided more data is available, recalling the prevalence of extrapolation for higher dimensions from Figure~\ref{fig:extrap_hist}.  Interpolation could be more prevalent if the data is chosen in a suitably intentional fashion (instead of randomly) or if the query domain of interest is largely contained in the convex hull of the data.  If the requisite latent space dimension is $\mathcal{O}\left(10^2\right)$ or larger, interpolation methods are unlikely to provide significant insight.}

\section{Discussion and Conclusions}
\label{sec:conclusion}

In this paper, we have proposed methods for utilizing interpolation techniques to validate scientific machine learning workflows and interpret {\sl how and why} various predictions were made during validation and testing.
Although interpolation techniques were traditionally believed to be prone to overfitting, we consistently observe similar performance between the best interpolation techniques and carefully trained ReLU MLPs.

In conclusion, a few rules of thumb from Section~\ref{sec:empirical-study} for using these methods include
\begin{itemize}
    \item TPS RBFs and GP interpolants perform extremely well for smooth problems;
    \item for nonsmooth problems, Delaunay interpolants appear to be more robust, and our revised Delaunay interpolation error bounds are more practically useful;
    \item based on the above, the smoothness of the underlying function is the most important property of the data set when determining interpolation or MLP regression performance;
    \item Delaunay interpolants consistently achieve similar performance to ReLU MLPs;
    \item other indicators of data set quality such as skew and spacing are less important, and for those methods that are affected, transforming/rescaling the data set (a standard practice) appears to address the issues; and
    \item extrapolation is a major hindrance to the tightness of our interpolation error bounds and also is a significant indicator of expected error.
\end{itemize}

Beyond these rules of thumb, we draw the following conclusions from our case-study in Section~\ref{sec:sciml-experiments}.

\begin{itemize}
    \item Interpolation methods, in addition to machine learning methods, should be a tool for scientific machine learning problems;
    \item error bounds for interpolation methods are useful diagnostic tools for understanding performance of these methods, and detecting potential issues in the training data and properties of the problem;
    \item geometric interpolation vs.~extrapolation is consistently a useful tool for judging approximation performance and can also be a warning for distribution shifts that lead to increased generalization error;
    \item interpolation methods can be leveraged within larger scientific machine learning pipelines, not as a replacement for, but alongside deep learning methods (such as representation learning);
    \item Delaunay interpolation and ReLU MLPs (when optimized) seem to achieve similar performance on real-world problems;
    \item as corroborated in other works, the right interpolation methods achieve similar performance to deep learning methods on real-world noisy problems, and their error bounds can be used to quantify this noise.
\end{itemize}

There remain opportunities to continue this work, studying the performance of interpolation techniques on a wider variety of scientific machine learning problems and considering additional options for our interpolation methods (particularly the GP methods) and alternative deep learning architectures beyond ReLU MLPs.

\paragraph{Reproduciblity}
Code used to generate the results appearing in this paper is available in a publicly available Github repository~\cite{interpMLrepo}.

\paragraph{Acknowledgments}
This work was performed under the auspices of the U.S.~Department of Energy by Lawrence Livermore National Laboratory under Contract DE--AC52--07NA27344 and the LLNL-LDRD Program under Project tracking No.\ 21--ERD--028.  
Release number LLNL--JRNL--846568.  
This work was also supported by the U.S.~Department of Energy, Office of Science, Advanced Scientific Computing Research, under Contract DE--AC02--06CH11357 and by award DOE--FOA--2493.

\bibliographystyle{elsarticle-num}
\bibliography{thc-refs,rm_refs}

\end{document}

%% file: preamble.tex
\usepackage{graphicx}
\usepackage[ruled,vlined,linesnumbered]{algorithm2e}

\usepackage{algpseudocode}
\usepackage{multicol}
\usepackage{subcaption}
\usepackage{epstopdf}
\usepackage{appendix}
\usepackage{enumitem}
\usepackage{hyperref}
\usepackage{color}
\usepackage{amsmath,amssymb,amsthm,amsopn}

\usepackage{cleveref}
\crefformat{equation}{(#2#1#3)}

\newcommand{\tail}{t}
\usepackage{bm} %

\numberwithin{theorem}{section}
\usepackage{tikz}
\usetikzlibrary{arrows.meta, positioning}